\documentclass{article}

\usepackage{arxiv}

\usepackage[utf8]{inputenc} 
\usepackage[T1]{fontenc}    
\usepackage{hyperref}       
\usepackage{url}            
\usepackage{booktabs}       
\usepackage{amsfonts}       
\usepackage{nicefrac}       
\usepackage{microtype}      
\usepackage{lipsum}
\usepackage{graphicx}
\usepackage{amsmath}
\graphicspath{ {./images/} }

\title{Privacy-Preserving Generative Models: A Comprehensive Survey}

\author{
 Debalina Padariya \\
  School of Computer Science and Informatics\\
  De Montfort University\\
  Leicester, United Kingdom \\
  \texttt{p2723446@my365.dmu.ac.uk} \\
   \And
 Isabel Wagner \\
  Department of Mathematics and Computer Science\\
  University of Basel\\
  Basel, Switzerland \\
  \texttt{isabel.wagner@unibas.ch} \\
  \And
 Aboozar Taherkhani \\
  School of Computer Science and Informatics\\
  De Montfort University\\
  Leicester, United Kingdom \\
  \texttt{aboozar.taherkhani@dmu.ac.uk} \\
  \And
 Eerke Boiten \\
  School of Computer Science and Informatics\\
  De Montfort University\\
  Leicester, United Kingdom \\
  \texttt{eerke.boiten@dmu.ac.uk} \\
}

\begin{document}
\maketitle
\begin{abstract}
Despite the generative model's groundbreaking success, the need to study its implications for privacy and utility becomes more urgent. Although many studies have demonstrated the privacy threats brought by GANs, no existing survey has systematically categorized the privacy and utility perspectives of GANs and VAEs. In this article, we comprehensively study privacy-preserving generative models, articulating the novel taxonomies for both privacy and utility metrics by analyzing 100 research publications. Finally, we discuss the current challenges and future research directions that help new researchers gain insight into the underlying concepts.
\end{abstract}

\section{\textbf{Introduction}}
 In recent years, the demand for generating valuable insights on applications while maintaining the privacy of individuals has increased. Synthetic data generation (SDG) is one of the emerging use cases of generative AI and has made significant progress as a privacy-enhancing technology \cite{bellovin2019privacy}. SDG aims to closely resemble real-world data, maintaining data privacy while preserving sufficient usefulness for analysis. Since the limitations of traditional de-identification methods are more evident, there is growing attention to SDG that captures the underlying distributions of real data. Although synthetic data holds great promise in multiple sectors, it is not a case of "Fake it till you make it" \cite{stadler_synthetic_2022}. The potential privacy attacks associated with generative models emerge as critical issues \cite{stadler_synthetic_2022,yoon_anonymization_2020}. Therefore, despite its promise for privacy-preserving data publishing, SDG requires adherence to regulations like GDPR, which restrict the collection and use of real data \cite{giomi_unified_2023}.
 \par Privacy-preserving generative models can address these challenges by releasing realistic synthetic samples  \cite{xie_differentially_2018, chen_gs-wgan_2020}. The popular approach is differential privacy (DP), which holds great promise for quantifying the privacy risk \cite{dwork_differential_2008}. However, DP is unlike having a "rich, calorie-free cake," and the trade-offs between privacy and utility are complex \cite{ganev_graphical_2023}.  There are many approaches for creating synthetic data with machine learning-based models, such as Generative Adversarial Networks (GANs), Variational Autoencoders (VAEs), statistical-based Gaussian Copula, transformer-based and agent-based models, or other ML-based methods \cite{lu_machine_2024}. This paper primarily focuses on the leading generative models, such as GANs and VAEs, which show remarkable performance in producing high-quality realistic synthetic samples \cite{arjovsky_wasserstein_2017}.
 \subsection{\textbf{Background}} \label{sec:background}
 Generative models have gained significant attention for unsupervised learning in a broad range of applications, such as image generation \cite{arjovsky_wasserstein_2017,radford2016unsupervisedrepresentationlearningdeep}, tabular data synthesis \cite{lee_invertible_2021,xu2019modeling}, time-series data generation \cite{frigerio_differentially_2019,beaulieu-jones_privacy-preserving_2019}, natural language processing \cite{sasada_differentially-private_2021}, and audio/video generation \cite{meyer_anonymizing_2023}. 
 Goodfellow et al. \cite{goodfellow_generative_2020} first proposed the GANs framework, which involves two neural networks competing in a minimax game. A generator network uses a random noise vector to generate new samples by learning the real data distribution. In contrast, a discriminator network works as a classification model to distinguish between real and generated samples. Many GAN variants have been proposed over the past few years, focusing on two objectives: improving training stability and deploying GANs for real-world applications \cite{radford2016unsupervisedrepresentationlearningdeep,mirza_conditional_2014,arjovsky_wasserstein_2017,gulrajani2017improved,karras_progressive_2018}.
 Knigma and Welling \cite{kingma2014stochastic} introduced VAEs, recognized as popular likelihood-based models. VAEs are associated with autoencoders comprising two interconnected networks: an encoder and a decoder. The encoder network takes the initial feature representation as input and transforms it into a dense representation.
Conversely, the decoder network is trained to reconstruct the original data from the encoded representation. A latent space or bottleneck consists of the compressed representation of the input data. Reconstruction loss measures how well a model reconstructs original data from its encoded form. The primary benefit of VAEs is their ability to control latent distributions, improving sample quality, but sometimes they can produce blurry outputs  \cite{mi_probe_2018}.
\par Despite the widespread success of generative models in various applications, several privacy threats have emerged as a significant concern \cite{chen_gan-leaks_2020,hayes_logan_2019,stadler_synthetic_2022} and privacy-preserving generative models have experienced rapid development in the past few years. DP provides a theoretical privacy guarantee through a noise-adding mechanism \cite{dwork_differential_2008}. In DP, the privacy parameters $\epsilon$ and $\delta$ quantify privacy loss by measuring the noise added to data. The pure DP mechanism, or $\epsilon$-DP, ensures that the output distribution remains almost identical for any two datasets that differ by at most one record. In approximate DP ($\epsilon,\delta$-DP), $\epsilon$ controls privacy, while $\delta$ represents the probability that the privacy guarantee might be violated. Differential privacy mechanisms have two key properties. First, composition refers to the joint distribution of the output of differentially private mechanisms that satisfy DP. Second, post-processing indicates that without additional knowledge of the private database, the function of the output of the differentially private algorithm can not be computed.
 \subsection{\textbf{Scope and Motivation}} 
\label{motivation}
 Some literature surveys have been published in the last few years on privacy aspects of generative models. Fan \cite{fan2020survey} provides a short review of existing approaches of differentially private GANs while emphasizing their evaluation metrics and application domains. However, this survey excludes the state-of-the-art privacy approaches in other generative models, e.g., VAEs, and only a limited number of differential privacy approaches applied to GANs. Additionally, privacy-preserving generative models have evolved over the past five years with effective noise-adding mechanisms in DP.
  Cai et al. \cite{cai_generative_2021} break down the GAN's privacy approaches into data and model-level while comparing the GAN-based security issues across different application areas. Although they cover different privacy metrics with a wide range of applications, the study lacks an in-depth analysis of them, covering only GAN-based privacy and security issues. Some surveys \cite{sun_adversarial_2023,zhang_generative_2023} briefly overview adversarial attacks in generative models while focusing only on attack-based privacy metrics, with a lack of guidance in identifying suitable metrics to maximize the benefits of generative models. Moreover, to our knowledge, no research has been conducted focusing on utility metrics in generative models. We address this gap by presenting our work, which seeks to provide the latest research around privacy and utility metrics with extended taxonomies and critically review the pros and cons of selecting particular metrics. We systematically analyze an exhaustive list of publications from top-tier conferences and digital sources. To achieve these objectives, the following research questions are addressed:

RQ1: What are the privacy attacks associated with generative models? 

RQ2: What are the approaches to quantifying privacy in generative models? 

RQ3: How can the utility of generative models be assessed?

RQ4: What are the open gaps and challenges in privacy-preserving generative models?
\subsection{\textbf{Contribution}}
Our contribution aims to comprehensively review the state-of-the-art privacy and utility metrics associated with generative models. This survey offers a deeper discussion of these concepts, categorization with several taxonomies, and insights into the open gaps and challenges with future research directions.
 The main contributions of this article are:
\begin{itemize}
\item  \textbf{Comprehensive Review:} This study comprehensively reviews the literature on privacy-preserving generative models. This is an in-depth analysis of over 100 research papers. To the best of our knowledge, this is the first work to provide a detailed comparison of both privacy and utility metrics. 
\item  \textbf{Taxonomies of Privacy and Utility Metrics:} There have been various privacy and utility metrics used in the current landscape of generative models — it is challenging and time-intensive to understand the broad picture of all works within this field. To offer readers a more accessible means of understanding existing studies, we provide novel taxonomies to categorize privacy and utility metrics of generative models. These taxonomies allow the reader to understand the similarities and differences of various metrics. We have provided taxonomies of privacy attacks in support of this and their adversarial assumptions against generative models. 
\item \textbf{Open Gaps and Challenges:} We elicit objectives for future research by assessing the state-of-the-art against the research questions posed. Based on the literature reviewed, we
shed light on the challenges yet to be solved and propose several promising future directions. 
\end{itemize}
The rest of the article is structured as follows:
Section \ref{sec:attacks} reviews the potential privacy attacks associated with generative models. Based on the proposed taxonomies, we review and discuss generative models' privacy and utility metrics in Section \ref{sec:privacy} and Section \ref{sec:utility}. We conclude our survey in Section \ref{sec:summary} by providing open challenges and future research directions. 
\section{\textbf{Attacks on Generative Models}} \label{sec:attacks}
The effects of attacks on the generation model are gaining more attention. Many studies have shown that generative models are vulnerable to attacks that break a model's integrity by disrupting the model's generation capacity or privacy by inferring sensitive information about the target model. This survey focuses on privacy attacks, where the vulnerability to attacks depends on the attack types and strategies (Section \ref{sec:attack-types}) and the attacker's assumption about the target model. It is the adversary's prior knowledge of the target model, such as training data distributions, model parameters, model structure, latent code, or generated samples. Based on the attacker's assumption, privacy attacks can be classified into two groups: Black-Box attacks and White-Box attacks. In full black-box attacks, the adversary can only access the synthetic samples from generative models \cite{stadler_synthetic_2022,hyeong_empirical_2022}, whereas in the partial black-box settings, the attacker may access partial knowledge of the target model, such as partial real training data or latent code of the target model \cite{hayes_logan_2019,chen_gan-leaks_2020}. On the other hand, in white-box settings, the attacker has some knowledge about the model, such as the training algorithm, model parameters, model architecture, or its original training data, such as partial or full training data distributions \cite{hu_model_2021,chen_gan-leaks_2020,park_evaluating_2021,mukherjee_privgan_2021}.
\begin{figure}[t]
    \centering
    \includegraphics[width=\linewidth]{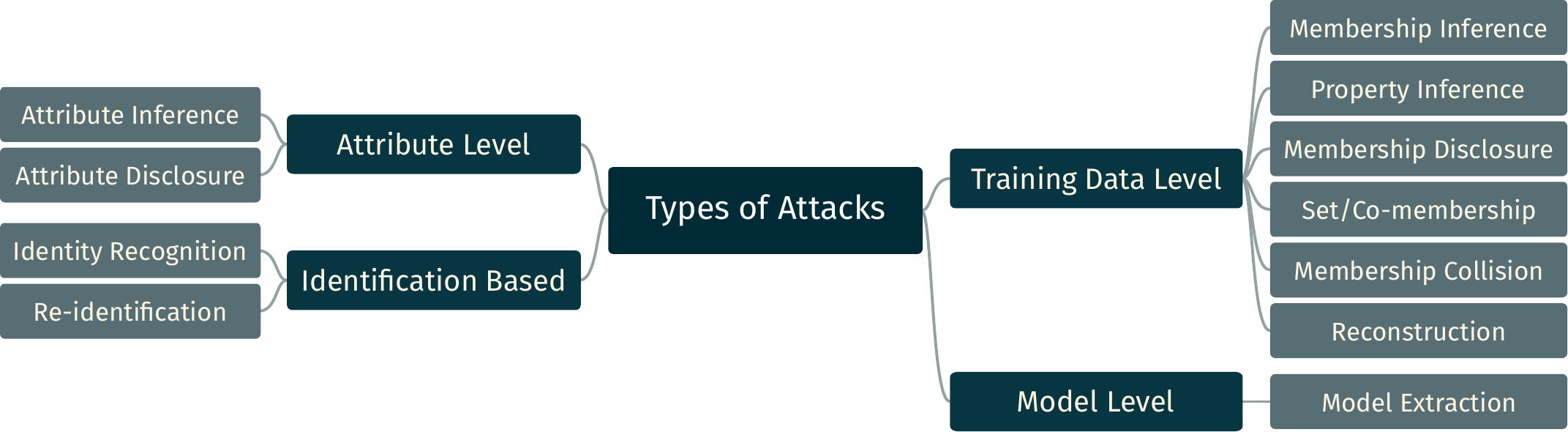}
    \caption{Types of Privacy Attacks in Generative Models}
    \label{fig:Types of Attacks}
\end{figure}
\subsection{\textbf{Types of Privacy Attacks}}  
\label{sec:attack-types}
The attacker aims to gain information that is not intended to be shared, such as information about the training data and attributes, the target model, or specific instances. The components of GANs and VAEs, such as the latent code, generator, discriminator, encoder, or decoder, are vulnerable to such attacks.
This section categorizes privacy attacks into four types (see Figure \ref{fig:Types of Attacks}): training data level, attribute level, model level, and identification-based.
\subsubsection{\textbf{Training Data Level}} 
At the training data level, the adversary aims to infer information about training data samples. For instance, if a model is trained on the genetic data of cancer patients, an adversary can infer individual cancer status or specific gene details of all patients.
Hence, we classify the training data-level privacy attacks into five categories: membership inference, property inference, membership disclosure, co-membership, and reconstruction.
\paragraph{Membership Inference Attacks ($MIA$)}
MIA is the most well-known privacy attack \cite{shokri2017membership}. 
In MIA, the attacker tries to determine whether an input sample $x$ is part of a training dataset $D$. A common MIA strategy is to build a local shadow model to mimic the target model and then manipulate the model. The idea is to gain more knowledge of the target model by building multiple shadow models. Because the output of the shadow models provides a prediction vector, it is useful to predict the membership in the training samples. The strategy behind shadow model training is related to overfitting, where a model works well on the training set but not on unseen data. Besides, shadow model training-based MIA against generative models \cite{xu_ganobfuscator_2019,frigerio_differentially_2019,park_data_2018,pan_privacy-enhanced_2023,yang_differential_2020,hu_tablegan-mca_2021}, variants of MIA also include distance-based MIA \cite{hilprecht2019monte, chen_gan-leaks_2020} and representation learning-based MIA \cite{zhang_membership_2022}.
\paragraph{Property Inference Attacks} 
These attacks attempt to infer a generic property of the training dataset from a target model \cite{zhou_property_2022}. 
The adversary queries the target model to achieve synthetic samples and then trains a property classifier to label these samples to infer the target property. 
In that case, the adversary can use shadow model training, using an auxiliary dataset to train the property classifier.
Property inference attacks can serve as the backbones for other attacks, i.e., membership inference attacks.
\paragraph{Membership Disclosure}
Through a membership disclosure attack, the attacker aims to identify an individual in the private dataset correctly, such as 
accessing a private patient record for model training.
Some researchers describe membership disclosure by accessing partial or complete sets of training samples \cite{goncalves_generation_2020, choi2017generating}. 
The intuition is that if a record is used to train the generative model, at least one synthetic sample is within a certain distance from the record. For instance, they use Hamming distance to compute the distance of each record to each sample from the synthetic dataset, where the attacker successfully identifies the records used for training at a smaller distance.
\paragraph{Set-membership/Co-membership Attacks} For set-membership, the adversary aims to identify whether a given set of records is used in the training set. 
Hilprecht et al. \cite{hilprecht2019monte} describe a set membership attack with the perspective of regulatory audits to prove data privacy violations, i.e., certain records are illegally used to train a generative model. To perform a set membership attack, the adversary holds some prior information, such as to which data source the samples belong, i.e., training or testing. First, this study launches single membership inference using these samples and then sorts the results with the distance function. Second, the top samples are recorded as a part of the training data. 
\par In another study, Liu et al. \cite{liu_performing_2019} perform co-membership attacks to identify whether all or none of the samples are used in the training set, e.g., multiple photos uploaded from a mobile. This study designs this attack against GANs and VAEs in two scenarios. First, the adversary mounts single membership inference on target samples by optimizing it using an attacker network. Second, the attacker network finds an optimal latent code using the query sample input. In that case, the adversary combines target samples to guide the attacker network in a co-membership attack, leveraging shared membership status to determine the final co-membership.
\paragraph{Membership Collision} This attack retrieves partial real training samples by accessing randomly sampled synthetic queries. Hu et al. \cite{hu_tablegan-mca_2021} propose membership collision attacks against GANs using synthesized tabular data. In tabular data, synthetic tables often overlap with the GAN's training samples, leading to a privacy violation if an adversary detects this intersection. They define this intersection as a membership collision. To perform this attack, this study uses a shadow model to train an attack model that predicts the collision probability of synthetic samples. Moreover, their findings show a positive relationship between membership collision and sample frequency, reflected as an indicator of membership collision.
\paragraph{Reconstruction Attacks} In this scenario, the adversary attempts to reconstruct one or more training samples. This is also called a model inversion attack, introduced by Fredrikson et al. \cite{fredrikson_model_2015}, where the adversary tries to recover the full data sample or its sensitive features. To investigate this attack, this study exploits a machine learning model's confidence score to analyze patterns and infer sensitive training data features. They use a face recognition system to reconstruct faces by training face classifiers using a linear regression model. Inspired by this approach, Li et al. \cite{li_privacy-preserving_2019} perform a reconstruction attack against GANs for a face recognition system. 
\subsubsection{\textbf{Attribute Level}} At the attribute level, the attacker tries to infer sensitive features of the attributes of individual training data, such as race, gender, or medical test results.
\paragraph{Attribute Inference Attacks} In this attack, the attacker attempts to predict the values of sensitive attributes by accessing the generative models. It is also referred to as linkage attacks, where the adversary tries to link specific sensitive attributes with a given data entry. Stadler et al. \cite{stadler_synthetic_2022} perform an attribute inference attack against GANs by developing an attribute inference game where the adversary can access the generated and partial real training samples. The idea is the adversary tries to predict an unknown attribute from known ones and then tries to infer the missing value via record linkage. If he succeeds in the linkage based on the known attributes, he reconstructs missing sensitive attributes; otherwise, the adversary uses the publicly released data to predict sensitive values. Inspired by this approach, several studies conduct attribute inference attacks against GANs \cite{houssiau_tapas_2022,kunar_dtgan_2021}.
\paragraph{Attribute Disclosure} This refers to the risk of an attacker correctly identifying the sensitive attributes of a target record based on the known attributes of real training samples. 
For example, the adversary can learn unknown attributes in a healthcare database by observing similar patient records. 
Some researchers describe attribute disclosure risk against GANs \cite{choi2017generating,goncalves_generation_2020}. To perform this, they consider $k$ - nearest neighboring distance, where the adversary can extract $k$ - nearest neighbor of the generated samples based on the known attributes achieved from the compromised record. 
\subsubsection{\textbf{Model Level}} At the model level, the adversary tries to take information from the target model and replicate the model. 
\paragraph{Model Extraction Attacks} In this scenario, the adversary attempts to replicate the full model by extracting information with the intention that the substitute model matches the target model's accuracy.  Hu et al. \cite{hu_model_2021} propose a model extraction attack against GANs. This study proposes an attack model for accuracy-based extraction aiming to steal the data distributions of the target model, i.e., generated and training data distributions. The strategy is querying the publicly released target model to gain the generated samples and then build a local copy of the target model. Additionally, to improve the attacker's effectiveness, they perform a fidelity-based extraction strategy, which involves two scenarios: accessing partial real-trained data in partial black-box settings and accessing the discriminator of the target model in white-box settings.
\subsubsection{\textbf{Identification Based}} In this scenario, the attacker attempts to recognize an individual identity based on their patterns, features, or characteristics. Alternatively, the attacker can re-identify individuals from an anonymized dataset.
\paragraph{Identity Recognition Attacks} Identity recognition identifies external entities' identity based on specific features or characteristics, i.e., face identification in face recognition systems. This occurs at the training data level, attribute, or feature level. For instance, pattern analysis in medical images happens at the training data level, while face identification focuses on specific attributes at the feature level. Chen et al. \cite{chen_vgan-based_2018} propose three threat scenarios to show the effectiveness of an identity recognition attacker for face recognition systems on VAEs and GANs. First, the adversary can target the unaltered training dataset and privacy-protected test dataset. Second, the attacker can access private training data to obtain the underlying ground truth identities and then train the identifier on training images, which consist of the same target identity as test images. In the last scenario, the attacker can retrieve the latent vector representations by controlling the encoder network.
\par Croft et al. \cite{croft_differentially_2022} explore identity recognition attacks for facial obfuscation against GANs by a parrot attack. During training, a parrot attack employs a neural network to classify identities by using labeled samples of obfuscated images in the training set, learning the identification pattern through the process.
In another study, Liu et al. \cite{liu_subverting_2021} describe identifying facial expressions in synthetic data against GANs. This study designs an adversarial approach, e.g., a steganography-based attack strategy, to understand how the adversary can extract sensitive attributes from input images, such as facial expressions. For instance, the adversary can extract sensitive information through the fully connected neural network layers to understand latent vectors, which helps to learn the feature representations of input in performing an identity recognition attack.
\paragraph{Re-identification Attacks} In this scenario, the attacker reveals the identity of a target record by linking it with some known information about the target record. Two types of identifying attributes increase the risk of re-identification. First, identifying attributes directly linked to an individual's identity, such as phone number or email address. Second is quasi-identifying attributes, which can be combined with other quasi-identities for re-identification attacks, i.e., date of birth, zip code, gender, etc. Yoon et al. \cite{yoon_anonymization_2020} analyze re-identification risk for patient data against GANs. For example, re-identification risk increases if a patient's date of hospital visits or medical procedures is combined with other attributes.
Gafni et al. \cite{gafni_live_2019} describe re-identification risk against GANs for video applications. In this study, the attacker can exploit the latent space representation to extract full information of an identity. In another study, Maxinmov et al. \cite{maximov_ciagan_2020} consider re-identification risk for people in video and image streams against GANs. They examine whether the attacker controls the pose preservation process or the temporal consistency, such as stability and continuity of video processing, as this makes GANs susceptible to re-identification risk. 
\section{\textbf{Privacy Metrics in Generative Models}} \label{sec:privacy}
\begin{figure}[t]
    \centering
    \includegraphics[width = \linewidth]{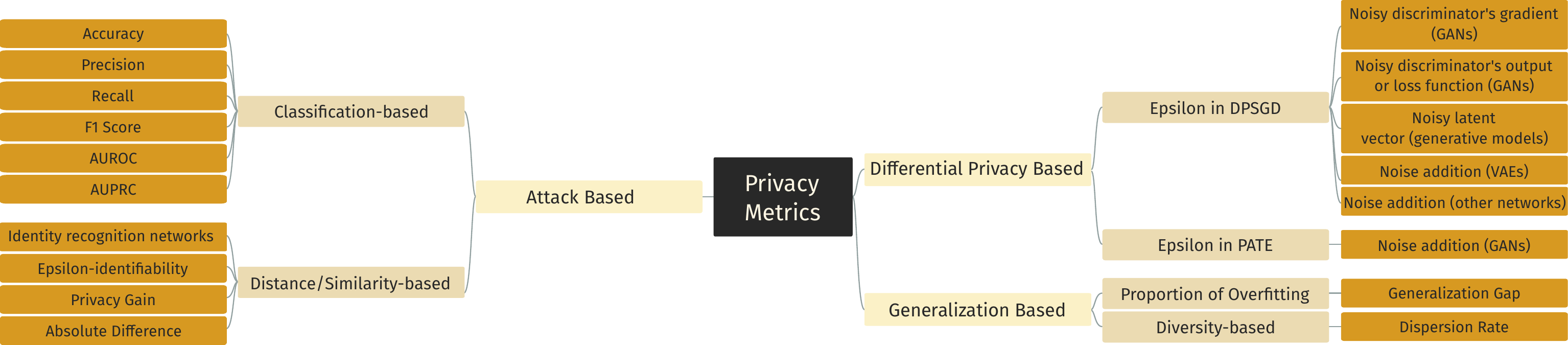}
    \caption{Taxonomy of Privacy Metrics in Generative Models}
    \label{fig: taxonomy of privacy metrics}
\end{figure}

Despite the generative model's success, emerging privacy threats have led researchers to improve privacy protection in this field. This survey explores the progress of privacy metrics by showcasing different privacy measures and the wide range of privacy guarantees established by researchers.
A generic approach is to use attack-based privacy metrics, where an attacker's success rate measures the privacy of the target generative model \cite{hayes_logan_2019,chen_gan-leaks_2020,stadler_synthetic_2022}. Alternatively, researchers have pointed out the poor generalization properties of generative models, where the proportion of overfitting can be a factor that measures information leakage \cite{chen_par-gan_2021}. The approaches based on differential privacy, which integrate noise-adding mechanisms,  provide insight into the overall privacy costs spent against generative models \cite{xie_differentially_2018,xu_ganobfuscator_2019,torkzadehmahani_dp-cgan_2019}. We propose a taxonomy of privacy metrics that allows categorizing different privacy measures, described in Figure \ref{fig: taxonomy of privacy metrics}. 
\subsection{\textbf{Attack based Metrics}}
In many approaches to privacy-preserving synthetic data, their privacy level is approached from the perspective of a specific attack. Many of these attacks return a value that may or may not be a valid result. Consequently, these attacks effectively present classification problems and their success; hence, the level of privacy can be measured based on confusion matrices. Others compare the distributions of original data and the outcomes of an attack on synthetic data. We summarize the metrics into two categories: classification-based and distance/similarity-based.
\subsubsection{\textbf{Classification-based Metrics}} This metric assesses how a classifier performs. The standard approach to evaluating the classifier's quality is via a confusion matrix. This matrix shows true positive, true negative, false positive, and false negative results for binary classification problems. Different metrics, such as accuracy, precision, recall,  and F1 score, are derived from the confusion matrix. Besides these, other classification metrics, such as AUROC (Area Under the Receiver Operating Characteristic Curve) or AUPRC (Area Under Precision-Recall Curve), are suitable for evaluating the classifier over all possible thresholds. 
\paragraph{Accuracy} This metric represents the percentage of accurately predicted instances compared to the total number of instances in a particular dataset.
For instance, the accuracy of membership inference attacks represents how many correct predictions the attacker makes within all observations. Many studies use this metric to measure the membership inference success rate against generative models \cite{hayes_logan_2019, hilprecht2019monte,park_evaluating_2021,mukherjee_privgan_2021,chen_vgan-based_2018,chen_gan-leaks_2020}. Some factors can influence the accuracy, such as knowledge about the target generative model, training dataset size or the number of training epochs. For instance, the partial black box-based MIA with limited auxiliary knowledge \cite{hayes_logan_2019} or knowledge about the latent code in white box settings \cite{chen_gan-leaks_2020,hilprecht2019monte} achieve higher accuracy than full black box-based MIA \cite{hayes_logan_2019}. Besides adversarial knowledge, some studies suggest that other factors, e.g., the size of training samples, can contribute to the success rate of MIA \cite{hayes_logan_2019,hilprecht2019monte,chen_gan-leaks_2020}. Since smaller training datasets lead to the model's overfitting, it is highly likely to increase membership inference accuracy.
\par The accuracy of a model extraction attack shows how well the attacker replicates the target model's behavior and performance \cite{hu_model_2021}.
The accuracy of an identity recognition attack indicates how effectively the attacker can correctly identify individuals within a given dataset or system. Liu et al. \cite{liu_subverting_2021} consider the discriminator's accuracy, where various factors can increase the discriminator's accuracy, such as network architectures, optimization techniques, training dataset size, or the number of training epochs.   
\paragraph{Precision/Recall} Precision measures the ratio of correct positive predictions (true positives) among all positive predictions in classifying the data instances. The precision of membership inference represents the fraction of records inferred that are indeed members of the training dataset. Alternatively, recall measures the ratio of correct positive instances (true positives) among all actual positive instances. For example, recall quantifies the fraction of the training dataset's members correctly inferred as members by the membership adversary. Goncalves et al. \cite{goncalves_generation_2020} use precision to measure membership disclosure and attribute disclosure success rate, where the adversary chooses a Hamming distance of each sample from the synthetic dataset to each sample in the training dataset. The distance of $0$ indicates high membership disclosure precision, indicating a high percentage of target records identified by the attacker. Conversely, a larger Hamming distance leads to higher recall, indicating the higher inclusion of target records in the training dataset. Regarding the success rate of these attacks, they report that the size of synthetic samples impacts recall with no effect on precision values.
\par Maximov et al. \cite{maximov_ciagan_2020} measure re-identification risk using a standard recall metric for image anonymization against GANs. This metric counts the proportion of nearest neighbor samples of the same class ranging from $0$ to $100$, where $100$ indicates a higher re-identification success rate. Additionally, a few works use a FaceNet identity recognition network to measure the success rate of identity recognition attacks against generative models \cite{gafni_live_2019,maximov_ciagan_2020}. The FaceNet network uses a true acceptance rate (TAR), the ratio of true positive identifications to false positive identifications, where the network achieves a higher identification success rate with a true acceptance rate of almost $0.99$.   
\paragraph{F1-Score} This is the harmonic mean of precision and recall. The F1 score of MIA represents a single measure of its effectiveness while identifying as many true members as possible and reducing the incorrectly identifying error of non-members. The F1 score is a better performance metric for imbalanced classes. Park et al. \cite{park_data_2018} use the F1 score to measure the success rate of membership inference and attribute disclosure with an imbalanced dataset. Since the F1 score balances precision and recall, a smaller F1 score indicates a declining proportion of correctly identifying members in membership inference or sensitive attributes of a target record in attribute disclosure. 
Besides, Hyeong et al. \cite{hyeong_empirical_2022} use the aggregation method of F1 score, i.e., Macro F1 score, calculated as the average of F1 score across all classes in multi-class classification problems to estimate the membership adversary's success rate.
\paragraph{AUROC} An AUROC visually represents a model's performance at different classification thresholds. This metric is used to evaluate binary classifiers, ranging from $0$ to $1$, where a value $0.5$ represents a random guessing. The AUROC of membership inference indicates how well the attacker can distinguish between the two classes: members and non-members. Some researchers have used this metric to measure the success rate of membership inference and co-membership attacks for GANs and VAEs \cite{liu_performing_2019,chen_par-gan_2021}. Their findings indicate that the attacker's AUROC score increases with the effective size of training samples and adversarial knowledge of the target model. Similarly, Park et al. \cite{park_data_2018} use the AUROC score to measure the MIA performance. Their findings suggest that improved quality of synthetic samples can produce more realistic cumulative distributions, resulting in a lower success rate of an attacker.
\paragraph{AUPRC} This metric is derived as the area under the precision-recall curve, which shows the trade-off between them across various thresholds. 
The AUPRC of MIA indicates the attacker's performance in identifying whether target individuals belong in the training dataset. 
AUPRC is more informative than AUROC for imbalanced data problems, i.e., where the negative class dominates, such as non-members, in membership inference. Since AUPRC measures how well the minority class is predicted, it gives less priority to false positives resulting from the majority class. 
Hu et al. \cite{hu_tablegan-mca_2021} use this metric to estimate the success rate for membership collision attacks against GANs, where a higher AUPRC indicates that the attacker accurately identifies the training sample. They also report that training data size and number of epochs can enhance the adversary's success rate. With the increasing training iterations, GANs learn more about the training data distribution, which enhances attack accuracy.
In another study, Bernau et al. \cite{bernau_assessing_2022} use an average precision metric to estimate the performance of reconstruction membership attacks against VAEs. The average precision represents the weighted mean of precisions obtained at each threshold.
\subsubsection{\textbf{Distance/Similarity-based Metrics}} These metrics are estimated between the distribution of the real and the synthetic samples, quantifying their dissimilarity. For instance, some studies have used these metrics to measure the privacy risks where the distribution of original data and the outcome of re-identification or identity recognition-based attacks on generative models estimate the level of privacy \cite{montenegro_privacy-preserving_2021,croft_differentially_2022}. Some identity recognition networks, such as FaceNet, ArcFace, and Siamese, particularly for facial recognition systems, measure the ability of the system to re-identify an individual across different datasets. Montenegro et al. \cite{montenegro_privacy-preserving_2021} have adopted a Siamese identity recognition network, which defines a maximum identity score, indicating the highest degree of similarity to predict the identity of privatized samples. Besides, some researchers have used FaceNet identification networks to estimate the re-identification success rate for face anonymization against GANs \cite{maximov_ciagan_2020,croft_differentially_2022}. The FaceNet network uses a similarity threshold by comparing the distance of two faces, where a value smaller than a threshold expresses the same identity \cite{schroff_facenet_2015}. Besides, few studies prefer another face recognition network, ArcFace, to measure the identification risks for facial images \cite{gafni_live_2019,tian_fairness_2022}. ArcFace uses a deep convolutional neural network (CNN), where a distance metric, i.e., cosine distance, is used to find a similarity score to identify a potential match \cite{deng_arcface_2019}. 
\par Yoon et al. \cite{yoon_anonymization_2020} propose a similarity-based metric, i.e., $\epsilon$-identifiability, to measure the re-identification risk against GANs for healthcare applications. The concept of identifiability is based on the fact that synthetic patient observations should differ enough from the original patient observations to ensure de-identification. They use a weighted Euclidean distance metric to measure the minimum distance between these two records. The $\epsilon$-identifiability defines a threshold with {($\epsilon \in [0,1]$)}, where $0$-identifiability represents a perfect de-identification, and $1$-identifiability implies perfect identification of a particular record. Besides, few studies have adopted the privacy gain metric, indicating a reduction in the adversary's advantage \cite{stadler_synthetic_2022, oprisanu_utility_2022}. This metric is estimated as the difference between the adversary's advantage over synthetic and original data, where a high privacy gain can protect all records from membership and attribute inference attacks. In another study, Zhou et al. \cite{zhou_property_2022} use the absolute difference metric to estimate the privacy risk for property inference attacks. The metric is calculated between training samples' inferred and ground truth proportions.
\subsection{\textbf{Generalization-based Metrics}}  Generalization refers to the model's ability to learn and predict the pattern of unseen data drawn from the same distribution of training datasets. Recent studies measure privacy by the proportion of overfitting in generative models and diversity-based metrics.
\subsubsection{\textbf{Proportion of Overfitting}} Overfitted models closely learn the training data and perform poorly on unseen data. If an overfitted model has a larger generalization gap, it makes this model vulnerable to membership inference. The generalization gap is the difference between model performance for training and hold-out data drawn from the same distributions. Chen et al. \cite{chen_par-gan_2021} measure the privacy of GANs by considering an extreme overfitting situation, which leads to a high generalization gap. The idea is to reduce the generalization gap to fix overfitting. This study uses multiple discriminators trained on disjoint partitions, where the generalization gap is measured by comparing the distribution of the discriminator's prediction scores between training and holdout data. Besides, they use two distance-based metrics, namely Wasserstein distance and Jensen Shannon divergence, to measure the similarity between two probability distributions, where similar distributions represent a smaller generalization gap. Their findings suggest that increasing disjoint partitions provide a similar distribution between training and holdout samples. Overall, multiple discriminators improve the generator's generalization ability, which can fix the overfitting problem.

\subsubsection{\textbf{Diversity-based metric}} A valuable property of a generative model is to produce diverse samples. Liu et al.\cite{liu_performing_2019} investigate how the diversity of generative models can serve as evidence of generalization using a geometric diversity metric, dispersion.
The generated and original data distribution dispersion scores in early training might be similar since the model may not generalize well. Later, when the model starts to generalize well, more diverse samples are produced, achieving a greater dispersion. However, Liu et al. \cite{liu_performing_2019} suggest that model generalization and diversity are inconsistent. 
\subsection{\textbf{Differential Privacy-based Metrics}}
Differential privacy (DP) is a well-known approach that promises to protect individual training samples in generative models. A parameter epsilon ($\epsilon$), known as privacy budget, controls the privacy level in DP. Many studies have adopted differential private training in generative models, where $\epsilon$ quantifies privacy by noise-adding mechanisms. The most common approach is to train the model using DPSGD (Differentially Private Stochastic Gradient Descent), adding Gaussian noise to the gradients during training. Others involve the PATE (Private Aggregation of Teacher Ensembles) mechanism, which is training distributed teacher models to transfer knowledge to generators.
 \subsubsection{\textbf{epsilon in DPSGD}} 
\begin{figure}[t]
    \centering
    \includegraphics[width=\linewidth]{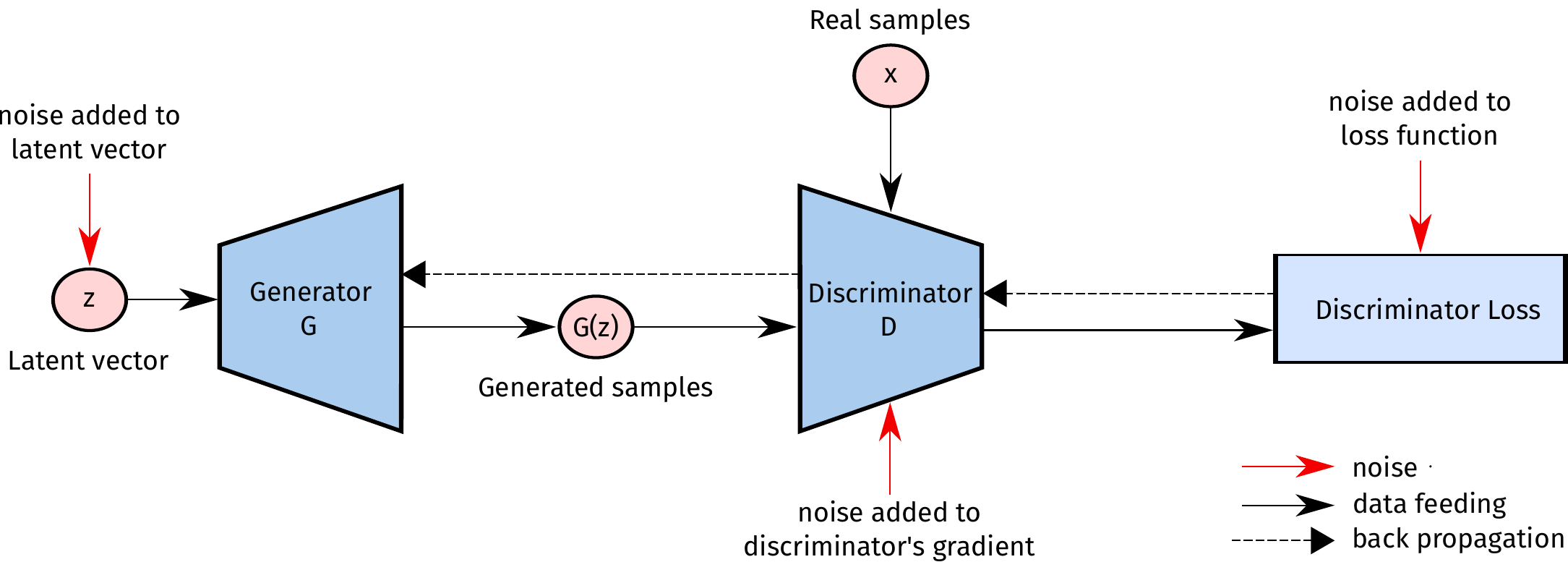}
    \caption{Noise Addition strategies in GANs}
    \label{fig:DPGAN}
\end{figure}

Abadi et al. \cite{abadi_deep_2016} introduced DPSGD for complex networks, ensuring a provable privacy guarantee in training samples. The DPSGD training procedure has two steps. First, gradients for model weights are computed from the loss of processed batch samples, and then computed gradients are clipped to control their sensitivity. The second step involves adding Gaussian noise to clipped gradients, updating the model, and estimating cumulative privacy loss with Moment Accountant. Finally, the training procedure terminates when the privacy budget is exhausted. Since its introduction, many researchers have used DPSGD to train GANs and VAEs using different noise-adding mechanisms, indicating effective use of privacy budget. Additionally, some studies work to improve convergence in learning algorithms by improving optimization strategies, which helps reduce privacy budget consumption.
 Hence, we have classified the noise perturbation strategies into four categories: noise addition to the discriminator's gradient in GANs, noise addition to the discriminator's output or loss function in GANs, noise addition techniques in VAEs, and noise addition to the latent vector in generative models. The various noise addition techniques in GANs are highlighted in Figure \ref{fig:DPGAN}. 
\paragraph{Noise addition to the discriminator's gradient in GANs} This is the most common strategy, where noise is perturbed to the discriminator's gradient. The idea is that the generator network trained by the differential private discriminator becomes differential private due to the post-processing property of DP. While choosing an appropriate privacy budget in DP remains unclear, researchers have shed light on how the sequence of noise addition affects privacy in practice. 
For example, carefully choosing the noise variance during training or using a strong composition theorem of DP for a tighter privacy budget. Yang et al. \cite{yang_differential_2020} have used
dynamic noise allocation strategies during the discriminator's training \cite{yang_differential_2020}. The idea is to add more noise during model training's early stages and gradually reduce it as parameters approach the optimal point to minimize the impact on convergence. This approach is further used in another study with a concentrated DP composition \cite{pan_privacy-enhanced_2023}. The study employs two adaptive noise allocation schemes, uniform and exponential decay, to dynamically choose the noise variance for the gradient. Uniform decay shows the uniform noise reduction to the minimum, whereas exponential decay indicates the noise scale decays exponentially. 
\par Xie et al. \cite{xie_differentially_2018} propose a privacy-preserving mechanism using the WGAN objective \cite{arjovsky_wasserstein_2017} by adding noise to the gradient of Wasserstein distance during the discriminator's training. This study uses a weight clipping strategy instead of gradient clipping like DPSGD, which helps the gradient automatically be bound by some constant to reduce privacy costs.
However, Xu et al. \cite{xu_ganobfuscator_2019} suggest that adding noise to Wasserstein distance gradients could lead to significant privacy loss due to slower convergence compared to regular GANs.
They introduce an adaptive gradient pruning strategy to track gradient magnitudes and dynamically adjust the pruning based on gradient updates.
Moreover, Alzantot et al. \cite{alzantot_differential_2019} use improved WGAN objectives \cite{gulrajani2017improved} to compute the gradient loss based on critic parameters from training data and add noise to the sum of clipped per-sample gradient. Further, 
Frigerio et al. \cite{frigerio_differentially_2019} propose clipping decay where the clipping bound decreases with each generator update. 
\par
\begin{figure}[t]
    \centering
    \includegraphics[width = \linewidth]{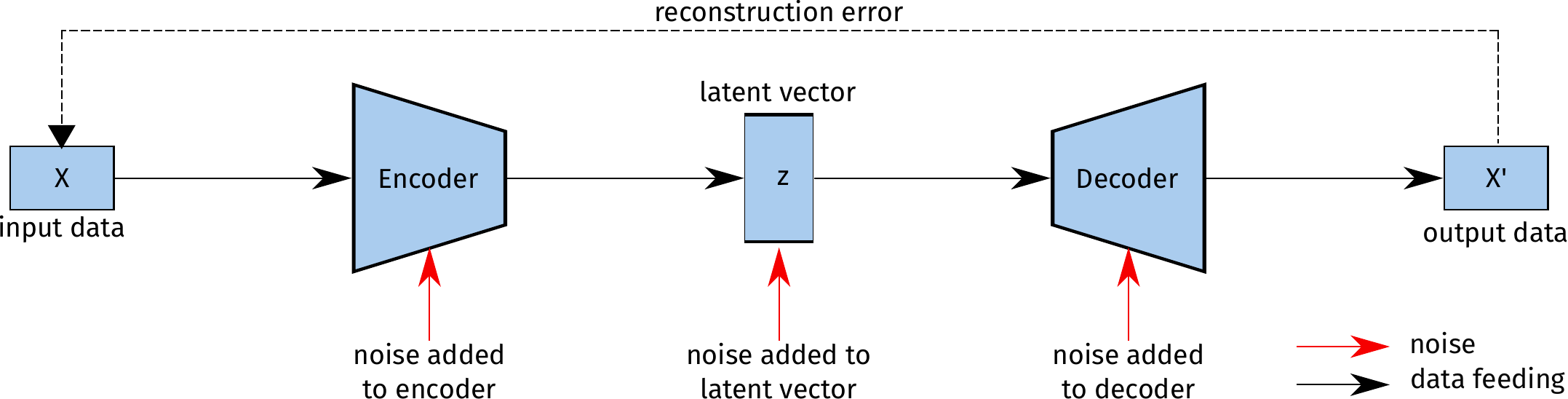}
    \caption{Noise Addition Strategies in VAEs}
    \label{fig: VAE}
\end{figure}

 Torkzadehmahani et al. \cite{torkzadehmahani_dp-cgan_2019} propose a privacy-preserving framework with CGAN objective \cite{mirza_conditional_2014} by clipping the gradients of discriminator loss on real and synthetic data separately. This study splits the discriminator's loss between real and synthetic data and then clips and sums the gradients of the two losses. Therefore, better control of the model's sensitivity can improve privacy. Also, they prefer the RDP mechanism over moment accounting, which offers a relaxed definition of DP that supports advanced composition theorems \cite{mironov_renyi_2017}.
In another study, Chen et al. \cite{chen_gs-wgan_2020} use an RDP accountant to train a discriminator privately with the improved WGAN objective \cite{gulrajani2017improved}. First, they adopt a subsampling strategy to reduce the privacy cost, where the whole dataset is subsampled into different subsets to train multiple discriminators.
Second, they selectively apply a noise-adding mechanism instead of gradient clipping like DPSGD, which helps to choose an optimal clipping value for a necessary subset of gradients, reducing noise effectively. This approach is further used in a few studies
\cite{zhao_ctab-gan_2024,kunar_dtgan_2021}. Besides, some researchers prefer Laplace noise to train the discriminator privately in GANs  \cite{imtiaz_synthetic_2021,ho_dp-gan_2021}. 
\paragraph{Noise addition to the discriminator's output or loss function in GANs} Han et al. \cite{han_differentially_2021} describe that existing frameworks that add noise to the gradient of discriminator's network during training are time-consuming and can decline training performance. This study focuses on efficient differential private GANs by adding noise to the discriminator's output or loss function. The idea is that by adding noise to the discriminator's loss function, the generator becomes differentially private by the post-processing property of DP.
They clip the discriminator's loss first and then add Gaussian noise to it. Ma et al. \cite{ma_rdp-gan_2023} propose a framework, RDP-GAN, where the Gaussian noise is added to the loss function of the discriminator for each iteration. They use an adaptive noise tuning strategy that dynamically adjusts the noise scale during training, allowing less noise addition into loss functions.
\paragraph{Noise addition techniques in VAEs}
Researchers have used different noise-adding mechanisms for VAEs, described in Figure \ref{fig: VAE}.
Acs et al. \cite{acs_differentially_2017} measure privacy using kernel-based clustering algorithms using a mixture of generative models, such as VAEs and RBMs (Restricted Boltzmann Machines). They describe that using mixture models requires less noise during training than single-model iterations. Also, this study partitions clusters and then trains the generative model on each cluster with DPSGD. Their findings suggest that using a mixture of generative models can require fewer training epochs than a single model, allowing less noise introduction.
In another study, Jiang et al. \cite{jiang_dp2-vae_2022} train the decoder privately using a subsampling strategy, where the dataset is randomly sampled and needs less noise than the full dataset.
Moreover, some studies estimate privacy with the two-phase training procedures, using GANs and autoencoders \cite{torfi_differentially_2022,tantipongpipat_differentially_2021}.
The idea is that when GANs are paired with autoencoders, the generator learns latent distributions accurately while the encoder reduces data dimensionality, allowing controlled noise injection.
For instance, Torfi et al. \cite{torfi_differentially_2022} use CGAN \cite{mirza_conditional_2014} and convolutional autoencoder, using RDP accountant for tighter privacy analysis while noise is added to the clipped gradient of the encoder and decoder. 

\paragraph{Noise addition to the latent vector in generative models} Some papers use private latent vectors to train generative models \cite{chen_generating_2022,takagi_p3gm_2021}. 
The intuition is that adding noise to the latent vector may reduce the privacy cost compared to injecting noise during model training. For instance, Chen et al. \cite{chen_generating_2022} design a differential private model inversion strategy to improve privacy over DPSGD-based GANs in two scenarios. First, a higher-dimensional GAN is trained with public data in the public domain, where no noise is introduced. Second, the resulting GAN is placed in the private domain, where a model inversion attack is performed to retrieve the latent vector. Then, a lower-dimensional GAN is trained with the collected latent vectors. Overall, privacy costs are reduced since no noise is introduced in the private domain, and the lower-dimensional GANs deal with less complex structures. 
In another study, Takagi et al. \cite{takagi_p3gm_2021} use an encoder-decoder network in a two-phase training procedure. 
First, the latent vector is privately trained with the DP-EM (Differentially Private Expectation Maximization) algorithm at the encoding phase. Then, the encoder is trained following the distributions of the latent variables. Second, at the decoding phase, the decoder is trained with the encoder, which improves training stability and increases noise resistance. 
\subsubsection{\textbf{epsilon in PATE}}
\begin{figure}[t]
    \centering
    \includegraphics[width = \linewidth]{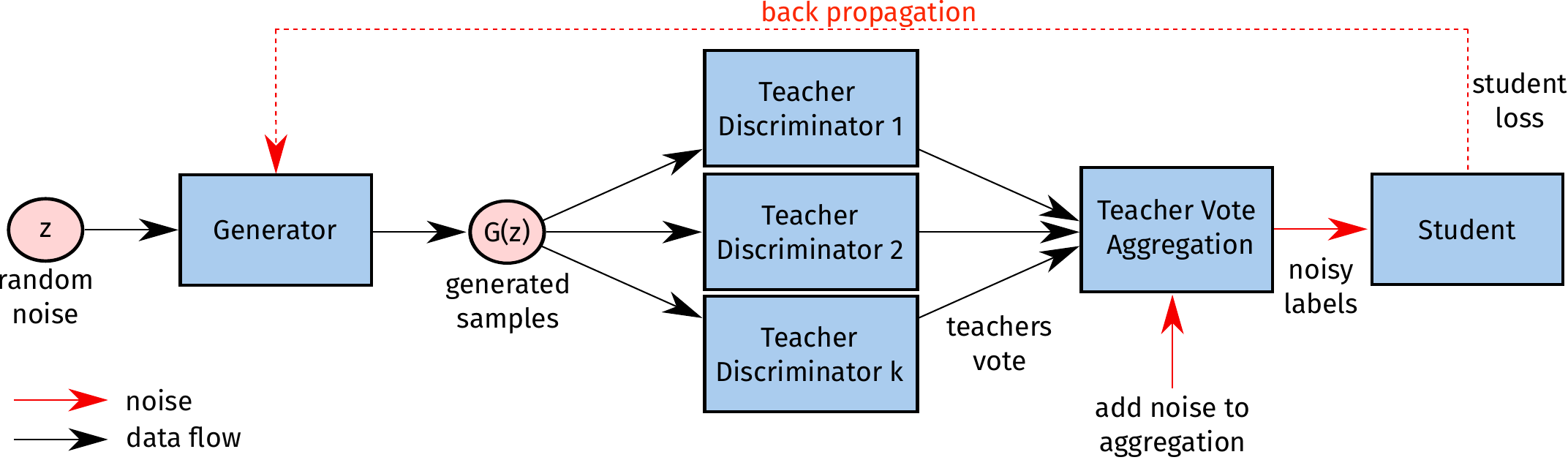}
    \caption{Noise Addition Techniques in GANs by PATE Mechanism \cite{jordon_pate-gan_2022}}
    \label{fig: pate}
\end{figure}

PATE is a generic approach providing DP guarantee in training samples, introduced by Papernot et al. \cite{papernot_scalable_2018}. In this method, the original sensitive dataset is partitioned into disjoint subsets, and then a teacher model is trained on each subset. The teacher model trains a student model while protecting the privacy of individual datasets. The main advantage is independent teacher training without restriction, where the predictions of all teachers' models are aggregated with noise-adding mechanisms to achieve the DP guarantee. Ultimately, knowledge transfer from the teacher models achieves a privacy-preserving student model. Inspired by this approach, some researchers have used PATE to train GANs privately. The noise addition technique through PATE mechanisms against GANs is described in Figure \ref{fig: pate}. 
\paragraph{Noise addition techniques in PATE-GANs}
 Jordon et al. \cite{jordon_pate-gan_2022} propose a framework, namely PATE-GAN, to train the GANs privately for multi-variate tabular data. The idea is to train distributed discriminators to transfer knowledge to the generator without an auxiliary public dataset. The teacher discriminators are trained individually on disjoint partitions of the training dataset, accessing only partial data. Then, the student discriminator is trained with generated samples labeled by the teachers using the PATE method. Accordingly, the generator is trained to minimize its loss function concerning the student discriminator.
Consequently, the student model can be trained privately without public data, allowing the generator to improve sample quality. 
\par In another study, Long et al. \cite{long_g-pate_2021} use a private gradient aggregation strategy that adds noise to the information flows from the teacher discriminators to the student generator.
Additionally, the aggregation mechanism includes gradient discretization and random projection to reduce the privacy budget used by each aggregation step. Random projection also reduces the dimensionality of gradient vectors on which the aggregation is performed. Furthermore, Zhuo et al. \cite{zhuo_pkdgan_2022} introduce a knowledge transfer framework, namely differentially private knowledge distillations that combine with GANs to transfer knowledge from teacher to student network. 
This study uses a privacy amplification strategy while adding noise to the sampled batch of distilled knowledge. Privacy amplification ensures that subsampling techniques are used efficiently for the privacy budget. 
\section{\textbf{Utility Metrics in Generative Models}} \label{sec:utility}
\begin{figure}[t]
    \centering
    \includegraphics[width = \linewidth]{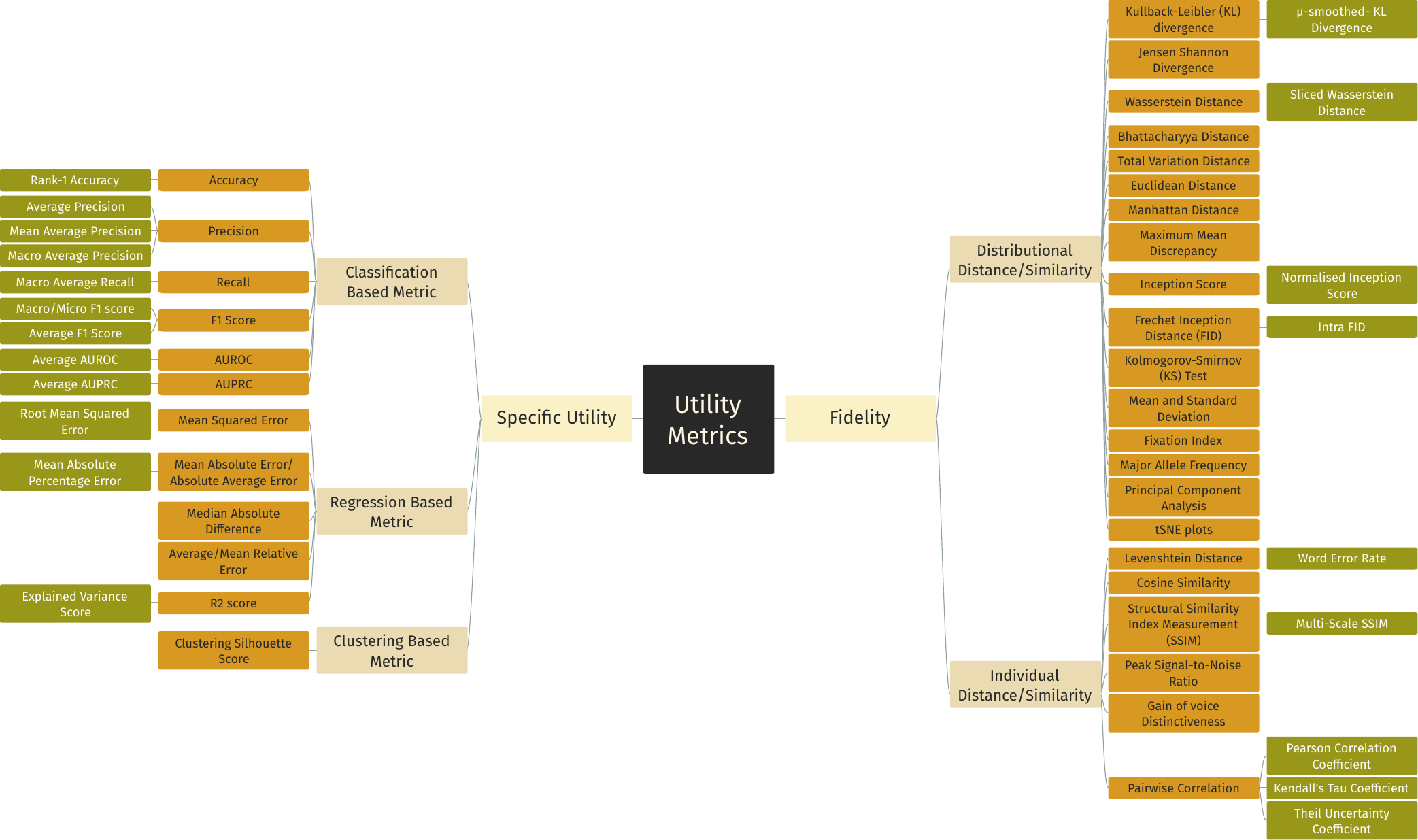}
    \caption{Taxonomy of Utility Metrics in Generative Models}
    \label{fig: taxonomy of utility metrics}
\end{figure}

In generative models, utility metrics measure and compare the efficacy of synthetic data. The intention is that synthetic data can complement or replace real data for training/inference purposes. High-quality synthetic data can be used for useful cases, while low quality generated data can lead to biases and inaccuracies.    
There are a variety of utility metrics depending on a wide range of applications. For instance, some users may be interested in finding the similarity between real and synthetic samples, while others pay attention to inferring information from the population during specific tasks. Based on this perspective, we propose a taxonomy of utility metrics, categorized into two parts: specific utility-based metrics and fidelity-based metrics, highlighted in Figure \ref{fig: taxonomy of utility metrics}.
\subsection{\textbf{Specific Utility-based Metrics}} 
These metrics measure the utility of synthetic data for particular analysis by comparing the results with real data, such as classification or regression tasks. Classification-based metrics measure the classifier's performance, whereas regression-based approaches assess the performance of predictive modeling. On the other hand, clustering-based metrics groups the data points into clusters based on their similarities or patterns.
\subsubsection{\textbf{Classification-based Metrics}}
These metrics measure how well synthetic data can complement real data in downstream classification tasks. The intuition is to assess the quality of synthetic data trained on machine learning models using different classifiers and test them on holdout data.
Researchers have used various classification-based metrics derived from the confusion matrix, such as accuracy, precision, recall, or F1-score, to measure the performance of synthetic data.
Some studies also use other classification-based metrics, such as AUROC or AUPRC, to evaluate the classifier's performance. 
\paragraph{Accuracy} This metric represents how well the generative model has learned the underlying distributions within the data, correctly predicting the class labels of their respective classes.  Xu et al. \cite{xu_fairgan_2018} use random samples from synthetic data to perform classification tasks against GANs, where the utility of generative models improves with the increased training samples. 
Regarding image-based datasets, researchers have measured the classification accuracy using different classifiers to distinguish whether the input images with a predefined label belong to the correct label \cite{ma_rdp-gan_2023,han_differentially_2021,chen_gs-wgan_2020}. Ganev et al. \cite{ganev_dp-sgd_2021} show that accuracy works better with balanced datasets than with imbalanced ones. Additionally, Frigerio et al. \cite{frigerio_differentially_2019} empirically show that achieving high accuracy depends on the strong correlations between different features. In another study, Zhuo et al. \cite{zhuo_pkdgan_2022} use Rank-1 accuracy, which refers to the percentage of predictions where the top prediction matches the ground-truth label to determine whether the ground-truth label equals the predicted class label with the highest probability. 
\paragraph{Precision/Recall} Researchers have used these metrics for binary and multi-class classification problems. For instance, Ganev et al. \cite{ganev_robin_2022} use precision and recall in binary and imbalanced multi-class classification problems with two classes: minority and majority classes/subgroups.
Because precision quantifies the number of correct positive predictions made, it calculates the accuracy for the minority class.
In another study, Kunar et al. \cite{kunar_dtgan_2021} use the Average Precision score, which uses a set of classifiers, whereas precision represents a specific classifier in a binary classification problem. Additionally, Zhuo et al. \cite{zhuo_pkdgan_2022} use the Mean Average Precision, which is the average of the Average Precision of each class. Moreover, some researchers have used Macro Average Precision, estimated by averaging the precision values calculated for each class without considering the class imbalance \cite{kotal_privetab_2022, chen_generating_2022}. These papers also use Macro Average Recall to show the model performance across all classes rather than focusing on classes with more samples. This is calculated as the sum of recall scores for all classes divided by the number of classes. 
\paragraph{F1-Score} Researchers have used this metric to show how synthetic data works on classification tasks for imbalanced classification problems against generative models \cite{tantipongpipat_differentially_2021,sasada_differentially-private_2021,kotal_privetab_2022,kunar_dtgan_2021,zhao_ctab-gan_2024}. Xu et al. \cite{xu2019modeling} use the Average F1-score, which is the weighted average of the class-wise F1 scores, where the number of samples available in that class determines the weights.
Some studies use Micro and Macro F1-score for multi-class classification problems against GANs \cite{chen_generating_2022,lee_invertible_2021,xu2019modeling,kim_oct-gan_2021}. The intuition is to calculate an overall F1 score for a dataset with more than one class. The Macro F1 score calculates the F1 score for each class and then averages them.
In contrast, the Micro F1 score calculates the total of True Positives (TP), False Positives (FP), and False Negatives (FN) for all classes and then calculates the F1 score from these totals. The Micro F1 score is useful for overall performance, whereas the Macro F1 score ensures all the classes perform well for imbalanced datasets.
\paragraph{AUROC} Researchers have used this metric in binary classification tasks to assess classifier performance on synthetic data, comparing how well the model distinguishes positive and negative classes with real data \cite{xie_differentially_2018,long_g-pate_2021,takagi_p3gm_2021,yoon_anonymization_2020}. Jordon et al. \cite{jordon_pate-gan_2022} use the Average AUROC score to show how well synthetic data captures the characteristics of real data against PATE-GANs for multiple models.
 \paragraph{AUPRC} Torfi et al. \cite{torfi_differentially_2022} suggest that AUPRC is more useful than AUROC for imbalanced classification problems. This is because AUPRC focuses on precision-recall trade-offs, particularly on finding the minority/rare true positive classes, which is crucial in imbalanced data. In another study, Jordon et al. \cite{jordon_pate-gan_2022} use an average AUPRC score across all classes, providing an overall performance measure for multi-class classification problems. 
 \subsubsection{\textbf{Regression-based Metrics}} 
\begin{figure}[t]
    \centering
    \includegraphics[width = \linewidth]{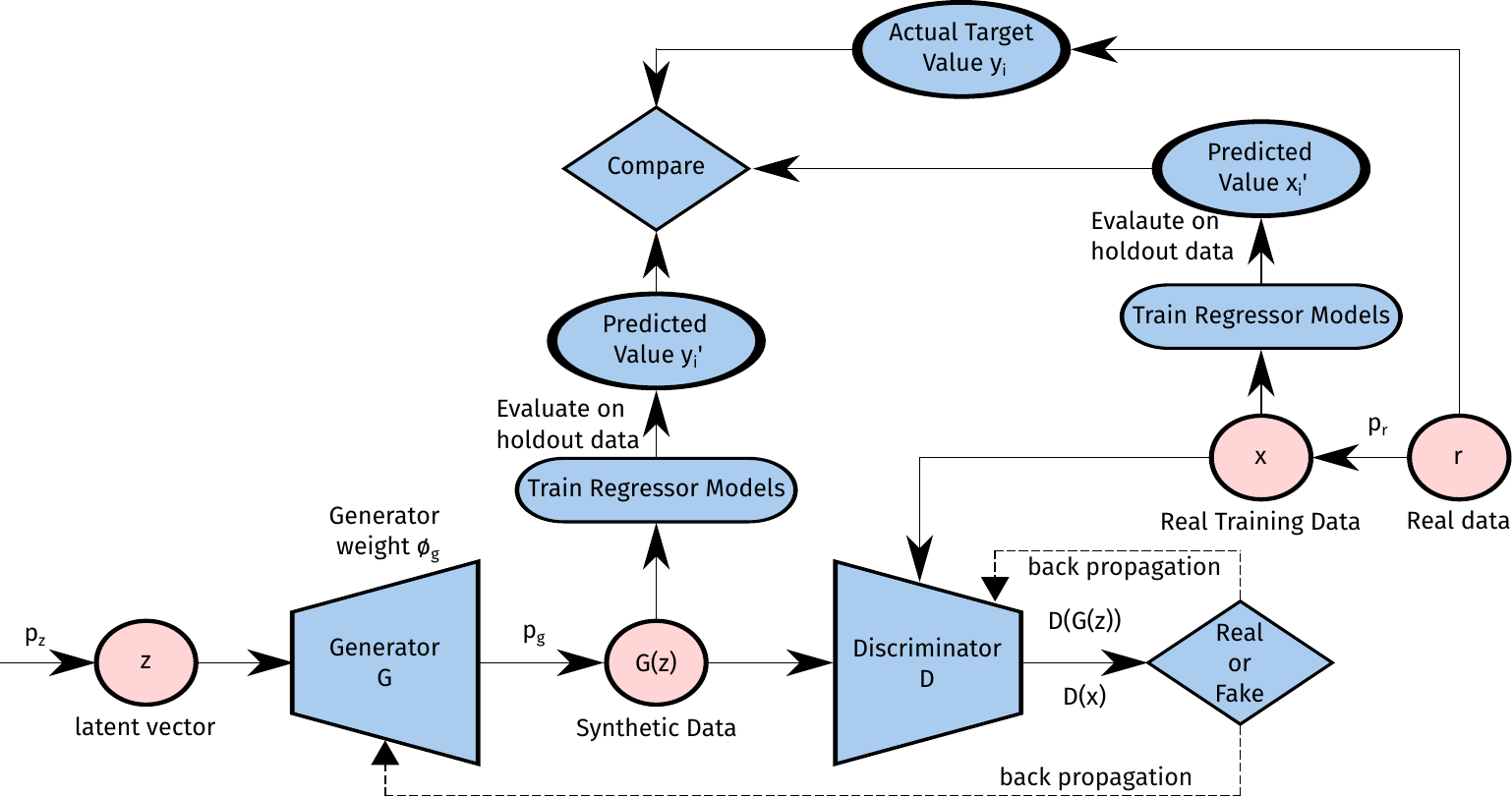}
    \caption{Evaluation Workflow of GAN-based Synthetic Data on Regression Tasks}
    \label{fig: Regression Tasks}
\end{figure}

The idea for calculating this metric is that the synthetic and real training data are trained on regression models and tested on hold-out data. An evaluation workflow on GAN-based synthetic data in regression analysis is shown in Figure \ref{fig: Regression Tasks}, where the actual target value of real data, i.e.,$y_i$, is compared with the predicted values of synthetic data, i.e., ${y_i}'$ and real training data, i.e., ${x_i}'$. These metrics predict a value by calculating an error score to show the model's predictive performance, such as mean squared error, mean absolute error, median absolute difference, or mean/average relative error. Besides, statistical regression-based metrics, i.e., R-squared ($R^2$) score, show the predictive performance of synthetic data compared to real data. The definition of all notations in regression tasks is shown in Table \ref{table:notation}.
\begin{table}[t]
\caption{Notation}
\label{table:notation}
\centering
\begin{tabular}{ll}
 \toprule
 Notation & Definition \\
 \midrule
 $G$ & Generator Network \\
 $\phi_g$ & Generator weight \\
 $D$ & Discriminator Network \\
 $z$ & Latent vector \\
 $p_z$ & Distribution of latent vector \\
 $r$ & Real data \\
 $x$ & Real training data \\
 $p_r$ & Distributions of real training data\\
 $G(z)$ & Synthetic data \\
 $p_g$ & Distribution of synthetic data\\
 $y_i$ & Actual target value  of the $i$-th record of real data \\
 ${y_i}'$ & Predicted value of the $i$-th record of synthetic data \\
 ${x_i}'$ & Predicted value of the $i$-th record of real training data \\
 $n$ &  Number of samples in the test dataset\\
 \bottomrule
\end{tabular}
\end{table}
 \paragraph{Mean Squared Error (MSE)/Root Mean Squared Error (RMSE)} The MSE of synthetic data is derived as the average of squared differences between actual target values produced by real data and predicted values generated by synthetic data. Similarly, Root Mean Squared Error (RMSE) is an extension of MSE, which calculates the squared root of MSE. RMSE is preferred over MSE because it provides results in the same units as the target variable, making it easier to understand the significance of errors. The MSE of synthetic data are defined in Equation \ref{eq:mse}, where all the notations are described in Table \ref{table:notation}:
\begin{equation}
    MSE = \frac{1}{n}\sum_{i=1}^n (y_i - {y_i}')^2 \label{eq:mse}
\end{equation}
 Lee et al. \cite{lee_invertible_2021} use MSE to demonstrate the quality of synthetic data for downstream regression tasks against GAN using tabular data. In another study, Ho et al. \cite{ho_dp-gan_2021} use RMSE to show the utility preservation of synthetic data using time-series datasets against GANs while applying this metric at each sampling time slot.
 \paragraph{Mean Absolute Error (MAE)/Mean Absolute Percentage Error (MAPE)} The MAE, also known as Absolute Average Error (AAE), of synthetic data, is estimated as the average of the absolute error values between actual target values on real data and predicted values on synthetic data. The MAPE of synthetic data measures the average absolute percentage error between the actual target values on real data and predicted values on synthetic data. MAPE is a scale-independent metric that can exceed $100\%$, allowing for better comparison of results across datasets with different units. The Equation \ref{eq:mae} represents the MAE of synthetic data:
\begin{equation}
    MAE = \frac{1}{n}\sum_{i=1}^n |y_i - {y_i}'| \label{eq:mae}
\end{equation}
 Some studies use these metrics against GANs where the synthetic data is trained using different regression models and tested on hold-out data to compare the predicted values with actual target values \cite{zhao_ctab-gan_2024,lee_invertible_2021, ma_rdp-gan_2023}.
 \paragraph{Median Absolute Difference (MAD)} This metric is calculated considering the median absolute differences between each data point and the overall median while comparing actual target values with predicted values. MAD of synthetic data is expressed in Equation \ref{eq:MAD}, where $\bar{y_i}=median(y_i)$:
\begin{equation}
    MAD = median(|y_i - \bar{y_i}|) \label{eq:MAD}
\end{equation}
Webster et al. \cite{webster_generating_2021} consider MAD to show how well the synthetic data captures the real data's features in vision tasks.
\paragraph{Mean Relative Error (MRE)} This metric measures the average errors, which are expressed as percentages. It focuses on the relative rather than absolute difference between actual and predicted values. This metric is also called Average Relative Error (ARE), shown in Equation \ref{eq:mre}:
 \begin{equation}
 MRE = \frac{1}{n} \sum_{i=0}^{n - 1} \frac{y_i - {y_i}'}{y_i} \times 100\%
 \label{eq:mre}
 \end{equation}
  Acs et al. \cite{acs_differentially_2017} use ARE to show the performance of synthetic data in regression tasks for time-series datasets. Besides, Gursoy et al. \cite{gursoy_utility-aware_2018} use ARE for privacy-preserving location trace problems while comparing actual and predicted answers for spatial counting queries.
 \paragraph{$R$-Squared ($R^2$) Score} This metric represents a dependent variable's variance ratio, which is further explained by an independent variable. In other words, the $R^2$ score assesses how well a regression model explains the variance in the data on which synthetic data is trained. This is also called Explained Variance, where a value of $0$ shows a poor fit, and $1$ indicates a perfect fit. The Equation \ref{eq:rs} represents the $R^2$ score for a linear regression model:
 \begin{equation}
 R^2 = \frac{SS_{regression}}{SS_{total}} \label{eq:rs}
 \end{equation}
 In equation \ref{eq:rs}, $SS_{regression}$ represents the variance explained by the model, which measures how well the regression model describes the data used for modeling. $SS_{total}$ shows the total variance, which measures the variation in the observed data. The total variance is measured by the sum of squares of the difference between predicted values and the actual target, described in Equation \ref{eq:TSS}:
 \begin{equation}
    SS_{total} = \frac{1}{n}\sum_{i=1}^n (y_i - {y_i}')^2 \label{eq:TSS}
\end{equation}
 Tantipongpipat et al. \cite{tantipongpipat_differentially_2021} report $R^2$-score using Lasso regression algorithms to measure the performance of synthetic data in regression analysis using mixed-type tabular data. Their empirical findings show that the synthetic data with continuous features has a negative $R^2$-score due to the lasso regression penalty term.
\subsubsection{\textbf{Clustering-based Metrics}}
These metrics measure the similarities between data based on their characteristics, grouping similar data objects into clusters while evaluating cluster distinctness using distance functions.
The distance function differs with variables such as boolean, categorical, ratio/ordinal, and interval-scaled, where weight is associated with the variables based on their applications. Some researchers have adopted clustering-based metrics to measure the quality and performance of synthetic samples compared to real ones in clustering algorithms, e.g., clustering silhouette score.
\paragraph{Clustering Silhouette Score (CSS)} This metric compares  
how close each cluster point is to the neighboring cluster's points. The range of CCS is $-1$ to $1$, where a higher score indicates the clusters are distinct. CSS is described in Equation \ref{eq:css}:
 \begin{equation}
 CSS = \frac{a-b}{max(b,a)} \label{eq:css}
 \end{equation}
In Equation \ref{eq:css}, $b$ shows the average distance between each point of a cluster, whereas $a$ indicates the average distance between all clusters. Kim et al. \cite{kim_oct-gan_2021} use CSS to evaluate the quality of generated samples using GANs for tabular datasets. They use the $k$-means clustering algorithm to group the data in real and generated samples and then calculate the silhouette score based on the resulting clusters. 
 \subsection{\textbf{Fidelity-based Metrics}}
 Fidelity measures how well the synthetic data in generative models matches the real data. This metric measures the quality of synthetic samples, where high fidelity shows their close distribution with real data. There are two common approaches to fidelity-based metrics: distributional or population level and individual sample level. The distributional or population-level fidelity metrics focus on the overall distributions of data points, i.e., how frequently each value appears in synthetic and real data distributions. Different ways, such as probability distributions, statistical tests, or histogram/visual representations, can compare overall distributions between real and synthetic data.
In contrast, the individual sample-level metrics compare the individual data points, calculating the distance or similarity between specific synthetic and real data values. Many researchers have measured the quality of synthetic samples using different datasets, such as images, tabular data, time series, graphs, audio/video, and genomics. 
\subsubsection{\textbf{Distributional Distance/Similarity-based Metrics}}
These metrics measure the difference between the entire distributions of real and synthetic data, focusing on the overall features or characteristics of the distributions rather than individual data points. Researchers have calculated the distributional distance or similarity by comparing probability distributions between synthetic and real data, such as Kullback-Leibler (KL) Divergence, Jensen-Shannon Divergence (JSD), Wasserstein Distance (WD), Bhattacharyya Distance (BD), Total Variation Distance (TVD), Euclidean Distance (ED), Manhattan Distance (MD), Maximum Mean Discrepancy (MMD), Frechet Inception Distance (FID) or Inception Score (IS). Besides, some papers have explored statistical similarity-based metrics, such as Mean and Standard deviation, Kolmogorov-Smirnov (KS) Test, Fixation Index (FI), or Major Allele Frequency (MAF), and visual representation-based metrics, such as Principal Component Analysis (PCA) or t-distributed Stochastic Neighbor Embedding (tSNE) plots, to compare real and synthetic data. 
\paragraph{Kullback-Leibler (KL) Divergence}
The KL divergence between two discrete probability distributions, such as $p_r$ for real data and $p_g$ for synthetic data, is defined in Equation \ref{eq:KL}:
\begin{equation}
    D_{KL}(p_r || p_g) = \sum_{x \sim X} p_r(x) * \log\left( \frac{p_r(x)}{p_g(x)} \right)
 \label{eq:KL}
\end{equation}
where $X$ is a random variable for all possible values to $x$. Additionally, $p_r(x)$ and $p_g(x)$ represent the probability of value $x$ occurring for the real and synthetic data distributions. Goncalves et al. \cite{goncalves_generation_2020} use KL-divergence to compare real and synthetic data distributions using tabular datasets, focusing on marginal probability mass functions (PMFs) for specific variables rather than entire datasets.
 In another study, Tantipongpipat et al. \cite{tantipongpipat_differentially_2021} have explored a variant of KL-divergence, i.e., $\mu$-smoothed KL- divergence between the real and synthetic distributions. This metric is used because standard KL-divergence struggles with zero probabilities in one distribution while $\mu$-smoothed KL-divergence offers a solution, represented in Equation \ref {eq:kl}:
\begin{equation}
    D_{\mu}(p_r \vert\vert p_g) = \sum_{x \sim X}( p_r(x) + \mu ) * log(\frac{p_r(x)+ \mu}{p_g(x)+ \mu})\label{eq:kl}
\end{equation}
 For example, in some cases where $p_r(x)$ is zero in Equation \ref{eq:KL}, while $p_g(x)$$>$$0$, KL-divergence becomes undefined.  
\paragraph{Jensen-Shannon Divergence (JSD)} The JSD between two probability distributions is expressed in Equation \ref{eq:JSD}, where $M$ is the mixture between real and synthetic data distributions.
$D(p_r \vert\vert M)$ and $D(p_g \vert\vert M)$ represent the KL Divergence from the real and synthetic data distribution to the mixture distribution:
\begin{equation}
    JSD(p_r \vert\vert p_g) = \frac{1}{2}*[D(p_r \vert\vert M) + D(p_g \vert\vert M)]   \label{eq:JSD}
\end{equation}
Kunar et al. \cite{kunar_dtgan_2021} use this metric to measure the difference between the probability mass distributions of categorical variables in the real and synthetic data. However, their empirical findings suggest that the JSD is unsuitable for continuous variables while real and synthetic data overlap. Moreover, Zhao et al. \cite{zhao_ctab-gan_2024} report that this metric is symmetric, which helps to interpret the experimental results easily by treating real and synthetic distributions equally. 
In another study, Gursoy et al. \cite{gursoy_utility-aware_2018} use this metric using location trace datasets to compute trip distributions of real and synthetic data, where JSD quantifies the trip error. 
\paragraph{Wasserstein Distance (WD)} This metric, known as Earth Mover Distance (EMD) or Wasserstein-$1$ Distance, measures the work needed to transform one distribution (synthetic data) to another (real data) by moving data points or mass.
This metric considers the absolute distance difference, where a lower distance shows similarity. Equation \ref{eq:WD} defines the Wasserstein Distance between two probability distributions, such as real and synthetic data distribution in a metric space $X$, denoted as $W_q(p_r, p_g)$, where $q$ is the exponent on the distance function used to calculate the cost of transporting mass between two probability distributions:
\begin{equation}
    W_q(p_r, p_g) = (\inf_{\gamma \in \Gamma(p_r, p_g)} \int_{X \times X} d(x, y)^q d\gamma(x, y))^{\frac{1}{q}}
    \label{eq:WD}
\end{equation}
where \ref{eq:WD}, $\Gamma(p_r, p_g)$ represents the set of all joint probability distributions, i.e., $\gamma$ on $X \times X$ whose marginals are $p_r$ and $p_g$. Additionally, $d(x,y)$ shows the distance between two points, such as $x$ and $y$ in the metric space $X$. Some studies have explored WD for continuous variables because min-max normalization, a key pre-processing step of WD, mitigates scale dependency issues when comparing distributions across features with varying scales, which is useful for continuous variables \cite{kunar_dtgan_2021,zhao_ctab-gan_2024}.
 In another study, Karras et al. \cite{karras_progressive_2018} use sliced Wasserestein distance, an alternative WD, to compare the probability distributions in high-dimensional settings. Because WD is computationally expensive for high-dimensional complex data, sliced Wasserstein distance overcomes this situation by approximating the high-dimensional distributions into multiple one-dimensional slices, calculating WD for each, and then the obtained distances are averaged.
\paragraph{Bhattacharyya Distance (BD)} This metric quantifies the similarity of two random distributions involving a Bhattacharyya coefficient. BD limits from $0$ to $1$, where a higher BD indicates the dissimilarity of two distributions. The BD for the two statistical distributions is expressed in Equation \ref{eq:cobha}: 
\begin{equation}
    D_B(p_r,p_g) = -ln(BC(p_r,p_g)) \label{eq:cobha}
\end{equation}
where
\begin{equation}
    BC(p_r,p_g) = \sum_{x \sim X} \sqrt{p_r(x)p_g(x)}  \label{eq:bha}
\end{equation}
In Equation \ref{eq:bha}, $BC(p_r, p_g)$ is the Bhattacharyya coefficient for discrete probability distributions, and $p_r(x)$ and $p_g(x)$ are the probabilities of event $x$ in two distributions. Bhattacharyya Distance translates the coefficient into a distance metric in Equation \ref{eq:cobha}. Lu et al. \cite{lu_poster_2017} apply this distance metric to compare real and synthetic data distributions using tabular datasets, where a larger Bhattacharyya distance indicates the dissimilarity between two distributions. 
\paragraph{Total Variation Distance (TVD)} This metric, sometimes called variational distance, compares the similarity of two probability distributions. TVD is expressed in Equation \ref{eq:TVD}, where the two probability distributions of real and synthetic data, such as $a_r$ and $a_g$ are defined in a sample space $X$, and $B$ is the subset of the sample space, representing a specific event:
\begin{equation}
    TVD(a_r, a_g) = \max_{B \subseteq X} |a_r(B) - a_g(B)|
 \label{eq:TVD}
\end{equation}
In Equation \ref{eq:TVD}, $a_r(B)$ and $a_g(B)$ represent the probabilities assigned to the subset $B$ by the two distributions. Thus, TVD is estimated as the maximum absolute difference over all possible subsets $B$ of $X$. Generally, this metric ranges from $0$ to $1$, where $0$ signifies perfect similarity and $1$ indicates maximum dissimilarity between two distributions. Takagi et al. \cite{takagi_p3gm_2021} calculate the average of TVD of all 2-way
marginals,i.e., relationships between pairs of variables of real and synthetic data. The intuition is that focusing on $2$-way marginals helps identify the data's dependencies and capture the core relationship while handling the complexity of high-dimensional data. Moreover, they consider the average TVD because it helps reduce sensitivity to outliers by considering all possible shifts of one distribution relative to the other. 
\paragraph{Euclidean Distance (ED)} This metric measures the direct or straight line distance between two points in Euclidean space, where the data points can be represented as feature vectors. Accordingly, when ED is applied to vectors, it calculates the overall difference between the corresponding elements of two vectors. Equation \ref{eq:ED} describes the Euclidean distance for two vectors, such as $X$ and $Y$, of real and synthetic data, and $n$ is the dimension of the feature vectors:
\begin{equation}
    ED(X,Y) = \sqrt{\sum_{i=1}^{n} (x_i - y_i)^2} \label{eq:ED}
\end{equation}
 Xu et al. \cite{xu_fairgan_2018} have used this metric to evaluate the closeness between synthetic and real datasets by considering the joint and conditional probabilities. They measure the Euclidean distance between the estimated probability vectors, namely probability mass function, derived from the sample space from the real and synthetic datasets. In another study, Hyeong et al. \cite{hyeong_empirical_2022} use the average ED (column-wise), focusing on the similarity of individual features between real and synthetic datasets.
\paragraph{Manhattan Distance (MD)} This metric, also known as $L1$ distance, calculates the distance between two points. MD considers the sum of horizontal and vertical distances between two data points by comparing the absolute differences in individual features of real and synthetic data. Equation \ref{eq:MD} defines the Manhattan Distance for two points $x$ and $y$, regarding real and synthetic data, where 
$n$ is the dimension of the feature vectors:
\begin{equation}
    MD(X,Y) = \sum_{i=1}^{n} |x_i - y_i| \label{eq:MD}
\end{equation}
Although this metric focuses on how much individual feature value differs between a real and synthetic data point, applying it to an entire distribution makes it a distributional distance/similarity measure. 
For instance, Lu et al. \cite{lu_poster_2017} consider this metric to compare the entire distribution of node degrees between the real and synthetic data using social network-based graph datasets. They apply this metric to compare the node attributes, where each node is associated with attribute vectors representing some characteristics. In another study, Li et al. \cite{li_privacy-preserving_2019} have used $L1$ distance to compare the occurrences of various activity types in real and synthetic process data, considering the overall distribution of activity frequencies.
\paragraph{Maximum Mean Discrepancy (MMD)}
This metric is a kernel-based statistical test that compares two probability distributions. It transforms the original data points into higher-dimensional feature spaces using kernel functions rather than comparing the original data points. MMD estimates the maximum difference between the average values of these transformed features across the two distributions, presented in Equation \ref{eq:MMD}:
\begin{equation}
   MMD(P, Q) = ||E_x \sim P[\phi(x)] - E_y \sim Q[\phi(y)]||_{H^2} \label{eq:MMD}
\end{equation}
where $E_x \sim P[\phi(x)$] and $E_y \sim Q[\phi(y)$] are the expected value of the transformed data point $\phi(x)$ and $\phi(y)$ under distribution $P$ and $Q$. $H^2$ is the squared norm in the Hilbert space $H$. Torfi et al. \cite{torfi_differentially_2022} use MMD to compare synthetic data to real data in a GAN framework because it is ideal for unsupervised settings and requires no labeled data. Also, it can effectively compare the distributions based on the inherent features captured by the kernel function.
\paragraph{Cosine Similarity} This metric estimates the similarity of two vectors in multi-dimensional space by deriving the cosine angle between the real and synthetic data vectors. Equation \ref{eq:COS} expresses the cosine similarity, where $A$ and $B$  represent the vectors of real and synthetic data points, and $\theta$ is the angle between vectors $A$ and $B$. Additionally, $\|{A}\| $ and $\|{B}\|$ are the magnitude of vector $A$ and $B$:
 \begin{equation}
     \cos(\theta) = \frac{{A} \cdot {B}}{\|{A}\| \|{B}\|}   \label{eq:COS}
 \end{equation}
 Sasada et al. \cite{sasada_differentially-private_2021} use this metric to compare the similarity between real and synthetic text. The idea is to focus on the directional similarity between two vectors using word frequencies in text, where high cosine similarity suggests similar overall distributions of words. In another study, Chen et al. \cite{chen_differentially_2021} have used this metric to assess synthetic sequences against real ones in financial applications. They use this metric because 
 it focuses on directional relationships between vectors rather than absolute positions or scales, which is ideal for time-series datasets.
\paragraph{Frechet Inception Distance (FID)} Heusel et al. \cite{heusel_gans_2018} first introduced FID to compare the distance between feature vectors calculated for real and synthetic images. This metric uses a pre-trained model, i.e., Inception $v3$, to retrieve the feature vectors for real and synthetic images. Then, the retrieved feature vectors are considered as datasets in a high-dimensional space. FID estimates the distance between these two feature vector distributions' statistical properties, such as mean and covariance. Equation \ref{eq:FID} describes the FID score considering real and synthetic image distributions:
\begin{equation}
    d^2 = || \mu_{\varphi}(p_r) - \mu_{\varphi}(p_g) ||^2 + Tr \left( \Sigma_{\varphi}(p_r) + \Sigma_{\varphi}(p_g) - 2 \sqrt{\Sigma_{\varphi}(p_r) \Sigma_{\varphi}(p_g)} \right) \label{eq:FID}
\end{equation}
where $\mu_{\varphi}(p_r)$ and $\mu_{\varphi}(p_g)$ are the mean of the feature vectors. $\Sigma_{\varphi}(p_r)$ and $\Sigma_{\varphi}(p_g)$ are the covariance matrices of the feature vectors for real and synthetic images. $d^2$ distance in squared units between two feature vector distributions, and $Tr$ is a trace operator estimated by summing the diagonal elements of the matrix. 
Long et al. \cite{long_g-pate_2021} use this evaluation metric to show how well synthetic images capture the statistical properties of real images, where a lower FID corresponds to better quality synthetic samples with more similar distributions with real data. Miyato et al. \cite{miyato_spectral_2018} proposed an intra-FID score, which addresses the limitation of standard FID for conditional GANs \cite{mirza_conditional_2014}. The intra-FID computes individual FID scores for each condition the cGAN is trained on and then calculates multiple FIDs corresponding to each conditional setting. 
Moreover, Karras et al. \cite{karras2019style} use the FID score to measure the quality of GAN-generated samples, where the FID score decreases while the training progresses. However, the FID score is reported as better early in training in another study \cite{bie_private_2023}.
\paragraph{Inception Score (IS)} This metric measures the quality of synthetic images to show how realistic they are compared to real images. Salimans et al. \cite{salimans_improved_2016} first proposed this metric for GANs, where a higher inception score indicates better-quality images. The inception score is calculated using a pre-trained image classification model, i.e., Inception $v3$ model, to predict the class probabilities, i.e., conditional class labels for synthetic images.
Equation \ref{eq:IS} defines IS of a generator $G$, where $z$ is the latent vector and $G(z)$ is the synthetic image:
\begin{equation}
IS(G) = exp(E_{x\sim G(z)}[{KL}(p(y|x)||p(y))]) \label{eq:IS}
\end{equation}
In Equation \ref{eq:IS}, $p(y|x)$ is the conditional class distribution for a given input image $x$, calculated by the inception-$v3$ model, and $p(y)$ is the marginal class distributions of synthetic images. Additionally, ${KL}(p(y|x)||p(y))$ is the KL-divergence of $p(y|x)$ and $p(y)$ and $E_x$ is the expectation of synthetic images. Chen et al. \cite{chen_differentially_2021} use this metric to evaluate the quality of GAN-generated images, where a larger IS score shows more diverse and better-quality synthetic images. In another study, Liu et al. \cite{liu_ppgan_2019} use a generated score, calculated by the mean of inception score, to evaluate the quality of generated images with real images. 
\paragraph{Mean and Standard Deviation} These metrics compare all data points between real and synthetic data. The mean represents the data distribution's central tendency. The standard deviation estimates the variation of the values in a dataset relative to the mean, described in 
Equation \ref{eq:MEAN} and \ref{eq:SD}:
\begin{equation}
    Mean = \sum_{i=1}^{n}\frac{x_i}{n} \label{eq:MEAN}
\end{equation}
\begin{equation}
    SD = \sqrt{\frac{\sum(x_i - x_{bar})^2}{n-1}} \label{eq:SD}
\end{equation}
In (\ref{eq:MEAN}) and (\ref{eq:SD}), $x_i$ is the $i$-th sample in a dataset, $n$ is the total number of samples, and ${x_{bar}}$ is the mean of the sample. Li et al. \cite{li_privacy-preserving_2019} use these statistical measures to compare the distance between synthetic and real sequences, where the number of activities in process data represents sequence length. 
\paragraph{Kolmogorov-Smirnov (KS) Test} This non-parametric statistical test compares the probability distributions of two datasets, i.e., real and synthetic data. Because this non-parametric test does not require knowledge of the underlying distributions, it is suitable for real-world data. The $KS$ test compares the two datasets' empirical cumulative distribution functions (ECDF), indicating the probability that a proportion of observations is less than or equal to a certain value. The idea is that the $KS$ test calculates a statistic, i.e., $D$, by comparing the ECDFs, representing the maximum absolute difference between two distributions, where a higher value shows a larger difference between two distributions. Equation \ref{eq:KS} defines the $KS$ test statistic $D$, where $F_P(x)$ and $F_Q(y)$ are the $ECDF$ of real and synthetic datasets at sample $x$ and $y$. Additionally, $F_P(x)$ at a point $x$ is calculated in Equation \ref{eq:ks}, where $n$ is the total number of samples in real dataset $P$:
\begin{equation}
    D = max | F_P(x) - F_Q(y) | \label{eq:KS}
\end{equation}
\begin{equation}
    F_P(x) = \frac{\text{Number of samples in } P \leq x}{n} \label{eq:ks}
\end{equation}
In the $KS$ test, a statistical reference table, the $KS$ table, provides critical $D$ values for various sample sizes at a significance level. A higher calculated value of the $KS$ test statistic $D$ than the critical $D$ value suggests a low $p$ value, indicating a significant difference between the two datasets' distributions. The $p$ is calculated by the test statistic $D$ and the sample sizes using statistical software (Python or $R$).
Oprisanu et al. \cite{oprisanu_utility_2022} use this non-parametric test for genomic datasets to compare the real and synthetic samples. In another study, Imtiaz et al. \cite{imtiaz_synthetic_2021} use $KS$ test on the samples taken from the real and synthetic distribution, where a higher $p$-value, indicates a higher probability that the samples come from the same distribution. 
\paragraph{Fixation Index (FI)} This population genetics measure shows how much genetic variation exists between subpopulations. The fixation index ranges from $0$ to $1$, where a value of $0$ indicates no genetic difference between subpopulations, and a value of $1$ suggests a complete genetic variance. The FD of synthetic data is defined as the ratio of variance among subpopulations to the total population variance, reflecting genetic differentiation.
Oprisanu et al. \cite{oprisanu_utility_2022} have adopted this population genetics measure to show how the groups of populations differ. In particular, they measure the variance in allele frequencies between populations to compare synthetic data with real data.
\paragraph{Major Allele Frequency (MAF)} This metric quantifies how frequently a most common allele occurs within a given population.
The MAF of real or synthetic data is calculated as the ratio of major allele counts to the total number of alleles, the count of major and minor alleles. Oprisanu et al. \cite{oprisanu_utility_2022} have considered this population genetics metric to show how well synthetic data complements real data against generative models. Additionally, they demonstrate that the sample size impacts MAF, where a larger sample size provides a more precise estimation of allele frequencies in a population.
\paragraph{Principal Component Analysis (PCA)} This metric is a linear dimensionality reduction technique used to compare the similarities in patterns or structures between real and synthetic data through visual representations. PCA is called a dimensional reduction technique. First, it identifies the most significant variance in data by focusing on the top principal components. Then, it reduces the dimensionality of the data while retaining most information. This metric follows three steps: standardization and covariance matrix, eigenvectors, and eigenvalues. First, a range of continuous initial variables is standardized, preventing features with larger scales from dominating the analysis and allowing PCA to identify the most significant directions of variance. Then, the covariance matrix is used to capture the correlations, i.e., linear relationships between features within data. Second, the eigenvectors and eigenvalues of the covariance matrix are derived to identify the principal components. Eigenvalues refer to the variance explained by each principal component, and eigenvectors show the directions of these principal components. Finally, a feature vector is constructed to determine the retained principal components. 
\par Yale et al. \cite{yale_generation_2020} use PCA to show the resemblance between synthetic and real data, whereas in another study, Xu et al. \cite{xu_ganobfuscator_2019} use an extension of PCA, i.e., $2$-way PCA, to analyze mixed-type complex datasets. The $2$-way PCA considers two-dimensional data, applying individually to each dimension. Their study reports that PCA provides promising results by producing detailed distributions in complex datasets.
\paragraph{t-distributed stochastic neighbor embedding (t-SNE)} This metric is an unsupervised non-linear dimensionality reduction technique that transforms high-dimensional data into a low-dimensional space for analysis and visualization. tSNE is used to visualize complex datasets into two or three dimensions, measuring the similarities by analyzing the underlying relationships. This minimizes the divergence between the original higher and lower-dimensional probability distributions. Then, the optimized lower dimension allows the visualization of the clusters and sub-clusters of similar data points.
Meyer et al. \cite{meyer_anonymizing_2023} have considered the tSNE measure to compare real and synthetic speech data, where they report that the t-SNE plots become similar with increasing training iterations. In another study, Tian et al. \cite{tian_fairness_2022} demonstrate a fairness protection performance in facial images, where they apply a tSNE measure to compare biased training data and privacy-preserving synthetic data distributions. 
\subsubsection{\textbf{Individual Distance/Similarity-based Metrics}}
These metrics measure how similar individual data points are between real and synthetic data. In other words, these metrics quantify the distance/similarity between a pair of data points to assess how well synthetic data captures the structure and relationship of real data. Researchers have adopted individual distance or similarity-based metrics, such as Levenshtein Distance, Structural Similarity Index Measure (SSIM), Peak Signal-to-Noise Ratio (PSNR), Gain of Voice Distinctiveness (GVD), or Pairwise Correlation Metrics.
\paragraph{Levenshtein Distance/Word Error Rate} This metric, also referred to as edit distance, measures the similarities between two sequences or strings, where the minimum number of operations, such as insertion, deletions, or substitutions, is required to transform from one sequence or string to another. Levenshtein Distance can be considered an individual distance and a distributional distance measure based on specific context. For example, Li et al. \cite{li_privacy-preserving_2019} have considered this metric for sequential/time-series datasets, applying this metric to the sequence variance to show the dissimilarity between sequences within a dataset. The intuition is to capture the overall variability within a set of sequences by comparing the Levenshtein distances between all pairs of sequences within the dataset. In another study, Meyer et al. \cite{meyer_anonymizing_2023} use Word Error Rate (WER) derived from Levenshtein Distance to compare the sequence of words of generated samples against the original sequence of words while focusing on the discrepancies between corresponding words.
\paragraph{Structural Similarity Index Measurement (SSIM)} This metric measures the similarity between real and synthetic images by evaluating an image's quality regarding luminance, contrast, and structure. This metric is calculated between the corresponding windows in the real and synthetic images at each scale, comparing the local patterns of pixel intensities. SSIM ranges from $-1$ to $1$, where a value of $1$ indicates perfect resemblance. SSIM is defined in Equation \ref{eq:SSIM}, where $a$ and $b$ are the windows of real and synthetic images, $\mu_a$ and $\mu_b$ are the pixel sample mean, $\sigma_a^2$ and $\sigma_b^2$ are the variance and $\sigma_{ab}$ is the covariance of $a$ and $b$:
\begin{equation}
\text{SSIM}(a,b)=\frac{{(2\mu_a\mu_b+x_1)(2\sigma_{ab}+y_2)}}{{(\mu_a^2+\mu_b^2+x_1)(\sigma_a^2+\sigma_b^2+y_2)}}
\label{eq:SSIM}
\end{equation}
where $x_1$ and $y_2$ are the constants added to stabilize the division when the mean and covariance are close to zero. Croft et al. \cite{croft_differentially_2022} have adopted this metric to measure the similarity of a facial obfuscated image to its real images in GANs. They calculate the average SSIM by comparing all obfuscated image pairs with identity classification accuracy. 
Researchers have also explored MS-SSIM, multi-scale variants of SSIM, which calculates a weighted average across scales for a more detailed image similarity assessment. Karras et al. \cite{karras_progressive_2018} use MS-SSIM in the training configurations of PPGANs while comparing synthetic images with real ones. They report that this metric is sensitive to high-frequency textures because PPGANs focus on broader structures over finer details, where this metric offers limited information.
\paragraph{Peak Signal-to-Noise Ratio (PSNR)} This metric quantifies the quality of synthetic images compared to real images, computed as the ratio between the maximum possible power of a signal/image, e.g., the maximum possible value of the data type and the background noise that affects the signal/image. The noise's power affects the data's fidelity, expressed in decibels (dB). PSNR is described in Equation \ref{eq:PSNR}, where $MAX$ is the maximum possible pixel values of real images and $MSE$ is the mean squared error:
\begin{equation}
    \text{PSNR} = 10 \times \log_{10} \left( \frac{{\text{MSE}}}{{\text{MAX}^2}} \right) \label{eq:PSNR}
\end{equation}
Wen et al. \cite{wen_identitydp_2022} have adopted this metric to measure the similarity of synthetic facial images with real images against GANs. They consider PSNR to show how well the intensity values of pixels in synthetic images correspond to those in real images, where a low PSNR suggests poor reconstruction quality.
\paragraph{Gain of Voice Distinctiveness (GVD)} This metric measures how closely a synthetic signal, i.e., the voice produced by a system, resembles the real signal, i.e., the human voice. GVD measures how well a voice processing system preserves the ability to distinguish between different speakers. This metric can be computed using different approaches, consisting of voice similarity metrics that compare the acoustic features. GVD values can be positive or negative, where a positive value indicates that the distinctiveness is preserved between real and synthetic voices. 
Meyer et al. \cite{meyer_anonymizing_2023} have applied this metric to evaluate the fidelity in a speaker anonymization process, which aims to hide the identity of a speaker by changing the voice during speech recordings. They use GVD to measure the difference in voice characteristics to show how well synthetic voices produced by GANs from various speakers remain distinct compared to the real speaker's voice.  
\paragraph{Pairwise Correlation Metrics} These metrics measure the linear relationship within real or synthetic data between all possible pairs of variables. In other words, it calculates the correlation for each pair of variables, resulting in a correlation matrix. Goncalves et al. \cite{goncalves_generation_2020} calculate pairwise correlation differences (PCD) to show how well the synthetic data captures the underlying correlation among the variables regarding real data. PCD is expressed in Equation \ref{eq:PCD}, where $A_r$ and $A_s$ are the real and synthetic data matrices: 
\begin{equation}
    PCD(A_r, A_s) = || \text{Corr}(A_r) - \text{Corr}(A_s) ||_F    \label{eq:PCD}
\end{equation}
In Equation \ref{eq:PCD}, $F$ is the Frobennius norm of the Pearson Correlation Matrices, calculated from real and synthetic datasets, where a smaller PCD suggests closer linear correlations across the variables. Chen et al. \cite{chen_generating_2022} have used the Pearson Correlation Coefficient (PCC) for time-series datasets that measure the linear relationship between two continuous variables. This metric ranges from $-1$ to $1$, where $1$ and $-1$ signify a linear relationship, and $0$ indicates no linear relationship. This metric is presented in Equation \ref{eq:PCC}, where $x_i$ and $y_i$ are the individual data points regarding real or synthetic data, $n$ is the number of data points, and $\bar{x}$ and $\bar{y}$ are the mean values of $x$ and $y$:
\begin{equation}
    PCC = \frac{\sum_{i=1}^{n} (x_i - \bar{x})(y_i - \bar{y})}{\sqrt{\sum_{i=1}^{n} (x_i - \bar{x})^2 \sum_{i=1}^{n} (y_i - \bar{y})^2}}    \label{eq:PCC}
\end{equation}
This metric is sensitive to outliers but performs well when the variables are normally distributed. 
In another study, Gursoy et al. \cite{gursoy_utility-aware_2018} apply the Kendall-Tau Coefficient (KTC) metric to compare the ranks within individual pairs. This metric is a rank-based correlation coefficient without relying on assumptions about the underlying distribution, such as normal or uniform distributions. This metric ranks data points instead of actual values to compare variables, where all possible pairs of data points are compared based on their ranks. This metric is represented in Equation \ref{eq:KTC}, where $\tau$ represents Kendall Tau Coefficient ranging from $-1$ to $+1$, and $n$ is the total number of data points in a dataset:
\begin{equation}
    \tau = \frac{{\text{{concordant pairs}} - \text{{discordant pairs}}}}{{n \times (n - 1) / 2}} \label{eq:KTC}
\end{equation}
In Equation \ref{eq:KTC}, concordant and discordant pairs indicate that the order of ranks of all pairs is the same and reversed regarding real or synthetic data. Moreover, Zhao et al. \cite{zhao_ctab-gan_2024} use the Theil uncertainty coefficient (TUC) to measure the correlation between any two categorical features within real and synthetic datasets. Because they deal with tabular datasets, this metric is well-suited for comparing distributions of categorical variables. This metric is typically used in information theory, quantifying the degree of uncertainty reduction about one variable given the knowledge of another variable, ranges from $0$ to $1$ and is described in Equation \ref{eq:TU} between two variables $X$ and $Y$:
\begin{equation}
    U(X|Y) = 1 - \frac{H(X|Y)}{H(X)} \label{eq:TU}
\end{equation}
In Equation \ref{eq:TU}, $H(X|Y)$ is the conditional entropy of 
$X$ given $Y$, measuring the uncertainty of $X$ given the value of 
$Y$ and $H(X)$ is the  unconditional entropy of $X$, measuring the uncertainty of $X$ without considering $Y$. This metric works well to capture both linear and non-linear relationships.  
\section{\textbf{Summary and Future Research Directions}} \label{sec:summary}
This article comprehensively reviews publications on the latest developments in privacy-preserving generative models, focusing on privacy attacks and metrics for privacy and utility. Starting with approximately 1200 papers, we identified 100 research publications for in-depth analysis, facilitating different aspects of potential threats, privacy, and utility metrics on generative models. We provided an explicit overview of privacy and utility metrics in generative models in Section \ref{sec:privacy} and \ref{sec:utility}. Despite much progress in this field in recent years, several open challenges remain for improving the existing approaches. This section highlights three key questions about the open gaps that need to be solved in future research. 
\subsection{\textbf{How to select privacy metrics in generative models considering the weaknesses and strengths in different metrics?}} \label{pm}
As demonstrated by several studies in Section \ref{sec:privacy}, each category of metrics has distinct benefits and implications. We observed that attack-based classification metrics primarily focus on data or model-level privacy by measuring the adversarial success rate, and they do not offer formal guarantees about the level of model privacy protection. Although researchers have used different classification-based metrics, there is no silver bullet since the choice of each metric depends on the particular requirements of a problem. For instance, single-score metrics, e.g., accuracy, are most common due to their effectiveness for balanced datasets and holistic approach to understanding how the model performs across all classes. However, for imbalanced and multi-class classification problems, AUROC or AUPRC is more appropriate.
Additionally, accuracy fails to provide insight into the model's prediction confidence and might overlook outliers. Similarly, outliers can influence precision metrics, potentially reducing True Positives. Besides, the recall fails to offer insights into the negative class instances. 
\par Selecting privacy metrics that reflect both the average and worst case is often recommended \cite{wagner_technical_2019}. However, distance/similarity-based attack metrics often focus on average-case performance and may not effectively address worst-case possibilities \cite{ganev_graphical_2023}. Even with a higher similarity, there could still be outliers in the data whose information remains vulnerable. Therefore, it may be beneficial to use robust distance-based metrics with less sensitive outliers, such as Median Absolute Deviation \cite{leys_detecting_2013} or combining similarity/distance-based metrics with other methods, e.g., clustering with outlier detection \cite{olukanmi_automatic_2022}.    
\par A wide range of studies rely on differential privacy-based metrics, which ensures a measurable privacy guarantee. However, the implications of DP in generative models are wide-ranging. First, in DP, adding more noise improves privacy and reduces task accuracy \cite{ganev_robin_2022,kunar_dtgan_2021,torfi_differentially_2022}. Second, determining the right privacy budget is complicated, and researchers have investigated the optimal selection of privacy budget to protect their models \cite{ganev_graphical_2023}. This is likely due to the complexity of setting the epsilon value for unpredictable datasets, which have high sensitivity and make it harder to find an appropriate epsilon.
Third, applying DP in healthcare can be challenging due to finite training samples \cite{yoon_anonymization_2020}. DP works well with large datasets but may be less effective with limited samples, as added noise can distort underlying data patterns. Finally, DP has an unforeseen side effect of reducing accuracy in underrepresented subgroups \cite{ganev_robin_2022}. While some researchers show DP has limited control over fairness \cite{bagdasaryan_differential_2019}, Cheng et al. \cite{cheng_can_2021} empirically explore that DP disproportionately increases the influence of majority subgroups, particularly in highly imbalanced datasets.
\par The generalization-based metrics improve model generalizability that can address overfitting problems to protect privacy in generative models \cite{chen_par-gan_2021}. Additionally, exploring the connection of the diversity of generative models with model generalization can be a promising research direction while measuring privacy \cite{liu_performing_2019}. Besides, some potential approaches, such as model pruning, knowledge distillation, and transfer learning, improve the model's generalizability to enhance its performance on specific tasks \cite{jin_prunings_2022,zhang_boosting_2023,ada_generalization_2022}. Model pruning reduces unnecessary weights, resulting in a smaller and faster model. Likewise, the transferring knowledge strategy in knowledge distillation improves the model's performance in adapting unseen data. Therefore, there is still scope for generalization-based metrics to improve privacy.
\subsection{\textbf{How to choose utility metrics for assessing generative models, considering their pros and cons?}}
As demonstrated by several studies in Section \ref{sec:utility}, the utility metrics are categorized into two parts. Section \ref{pm} already discusses several issues regarding classification-based metrics, which apply similarly to such metrics when used for utility. Besides, existing studies have used four categories of regression-based metrics: scale-dependent measures, such as MSE, RMSE, MAPE, and MAE; measures based on relative error, e.g., MRE; scaled error, e.g., MAD and goodness-of-fit metric, e.g., $R$-Squared score. However, there are no ideal metrics that satisfy all conditions. For instance, scale-dependent metrics, such as MSE or RMSE, are suitable when data is normally distributed; however, these are affected by outliers. For example, RMSE focuses on large errors, which might not be as important for capturing overall trends in time series data \cite{armstrong_error_1992}. MAE and MAD, which consider absolute deviations, can better handle outliers in such cases.
Moreover, MAPE, expressed as a percentage, makes comparing forecasts across time series with different scales challenging \cite{foss_simulation_2003}. Also, its sensitivity to zero values can be problematic in time series datasets with frequent zeros or very low values. The $R$-Squared score works well for linear relationships; however, it is unsuitable for assessing predictive accuracy in numerical data, as it focuses on error direction rather than magnitude \cite{li_assessing_2017}. Therefore, researchers should consider suitable regression metrics depending on some specific context, such as the type of errors, outliers, or data distributions. 
\begin{table*}
  \caption{Fidelity Metrics Based-on Data Types}
  \label{table:datatypes}
  \begin{tabular}{p{4cm} p{4cm} p{4cm}}
  \toprule
  Data Types & Utility Metrics & References \\
 \midrule
  Image & IS, FID, PSNR, SSIM, KL-Divergence, JSD, WD  & \cite{mugunthan_dpd-infogan_2021,wen_identitydp_2022,kuang_unnoticeable_2021,xu_ganobfuscator_2019,webster_generating_2021,maximov_ciagan_2020,azadmanesh_white-box_2021,zhou_property_2022,long_g-pate_2021,chen_gan-leaks_2020,croft_differentially_2022,bernau_assessing_2022,karras_progressive_2018,yang_differential_2020} \\
 Tabular & PCD, BD, JSD, TUC, KS Test, ED, WD  & \cite{lu_poster_2017,zhao_ctab-gan_2024,imtiaz_synthetic_2021,kunar_dtgan_2021,goncalves_generation_2020,hyeong_empirical_2022,tantipongpipat_differentially_2021} \\
 \addlinespace
 Time-Series & MMD, PCC, KTC & \cite{gursoy_utility-aware_2018,chen_generating_2022,beaulieu-jones_privacy-preserving_2019} \\
 Audio/video & WER, GVD  & \cite{meyer_anonymizing_2023} \\
 \addlinespace
 Genomics &  FI, MAF, KS Test & \cite{oprisanu_utility_2022}\\
  \bottomrule
\end{tabular}
\end{table*}
\par Regarding fidelity-based metrics, it remains unclear when their scores are meaningful and when they may be misinterpreted.
For example, while PSNR measures image fidelity in preserving visual details \cite{wen_identitydp_2022}, its relevance to audio fidelity for preserving sound quality is unclear. Similarly, pairwise correlation-based metrics are primarily applied for tabular and time-series datasets \cite{zhao_ctab-gan_2024}; however, their efficacy for image datasets remains uncertain. This is likely because correlation-based metrics capture linear relationships, whereas image datasets often involve more complex relationships. Cosine similarity, used in text recognition \cite{sasada_differentially-private_2021}, measures vector direction rather than magnitude, which could benefit other domains, i.e., time series analysis \cite{wong_visualizing_2019,dong_cosine_2006}. Likewise, Maximum Mean Discrepancy (MMD), used for measuring distributional similarity for time-series datasets \cite{torfi_differentially_2022}, can also be applied to other domains, such as tabular, image, or text due to its ability to capture complex relationships and flexibility of its kernel function \cite{zhang_understanding_2022,bista_supmmd_2020}.
\par In some scenarios, the data types can impact the choice of fidelity-based utility metrics, described in Table \ref{table:datatypes}. For instance, some standard metrics quantify the visual similarity of image datasets, such as IS, FID, SSIM, and PSNR. Researchers have used these because they provide quantitative measures, focusing on the perceptual similarity or human perception that assesses how well the synthetic images capture the underlying features of real images. However, human assessment tends to be biased toward the visual quality of generated samples, where evaluations can be inconsistent and hard to compare.
Similarly, some metrics estimate different aspects of tabular datasets, such as numerical or categorical features, relationships, or distributions within data. For example, pairwise correlation considers linear relationships in numerical data, whereas Bhattacharyya Distance (BD) compares the probability distributions of categorical or binary features. 
Moreover, the Pearson Correlation Coefficient (PCC) is used in time series to estimate linear dependence, which makes it easy to interpret the trends.
Maximum Mean Discrepancy (MMD) detects distributional differences, while the non-parametric Kendall-Tau Coefficient is robust to outliers, making it ideal for time-series analysis. Further, some metrics consider audio quality assessment, i.e., Gain of Voice Distinctiveness (GVD) or different aspects of genomic analysis, such as the Fixation Index (FD), Major Allele Frequency (MAF), or Kolmogorov-Smirnov (KS) test.
\par It is crucial to consider the target audience when choosing metrics. Simple metrics are easy to understand for non-experts, while visual comparisons are more attractive. Individual-based metrics provide limited scope, but merging them with distributional-based metrics can provide a detailed analysis. However, implementing them in a particular scenario can pose challenges due to specific metrics features that are not predicted, such as sensitivity to outliers, measurement bias, issues in data quality, or incompatibility in some contexts. Because there is no one-size-fits-all solution, researchers should select metrics based on their specific objectives.
\subsection{\textbf{What is the importance of fairness metrics in generative models?}}
Generative models-based synthetic data can inherit biases from the algorithms used to learn from real-world training data \cite{chen_would_2024}. Some researchers have explored bias for facial applications, such as inaccuracies from race or gender bias \cite{tian_fairness_2022} or reduced utility in text recognition due to repetitive words \cite{sasada_differentially-private_2021}. Fairness in synthetic data involves recognizing and rectifying biases in training data and algorithms to avoid discriminating against certain groups. While researchers have explored the potential of fairness metrics to handle bias in synthetic data \cite{tian_fairness_2022}, Cheng et al. \cite{cheng_can_2021} have analyzed the significance of fairness metrics in downstream performance. In downstream tasks, researchers rely on application-specific metrics. However, depending on these metrics can lead to overfitting specific tasks while failing to capture the broader distribution of real data. Fairness metrics can help capture underlying distributions across various groups. Therefore, there is still room for significant contribution to fairness in generative models, such as fairness metrics \cite{tian_fairness_2022}, debiasing techniques \cite{draghi_identifying_2024}, or counterfactual fairness \cite{abroshan_counterfactual_2022}.

\section{Acknowledgments}
This work was supported by the Alan Turing Institute under the Turing/Accenture strategic partnership grant R-AST-040.

\bibliographystyle{plain}


\end{document}